\documentclass[journal]{IEEEtran}
\usepackage{times}
\usepackage{soul}
\usepackage{url}
\usepackage{xr-hyper}
\usepackage[hidelinks]{hyperref}
\usepackage[utf8]{inputenc}
\usepackage[small]{caption}
\usepackage{graphicx}
\usepackage{amsmath}
\usepackage{amsthm}
\usepackage{booktabs}
\usepackage{soul}
\usepackage{amsmath,amssymb}
\usepackage{subcaption}
\usepackage{xcolor}
\usepackage{algorithm}
\usepackage{algorithmicx}
\usepackage[noend]{algpseudocode}
\usepackage{makecell}
\usepackage{bbm}
\usepackage{pgfplots}
\usepackage{float}

\newtheorem{theorem}{Theorem}
\newtheorem{lemma}{Lemma}
\newtheorem{define}{Definition}
\newtheorem{assumption}{Assumption}
\newtheorem{proposition}{Proposition}
\newtheorem{corollary}{Corollary}
\newcommand{\infoset}{infoset}

\newcommand{\D}{\Delta}

\DeclareMathOperator*{\argmax}{argmax}

\DeclareMathOperator{\BR}{BR}

\makeatletter
\newcommand{\printfnsymbol}[1]{%
  \textsuperscript{\@fnsymbol{#1}}%
}
\newcommand{\printsymbol}[1]{%
  {\@fnsymbol{#1}}%
}
\makeatother

\begin{document}

\title{Model-Free Neural Counterfactual Regret Minimization with Bootstrap Learning}

\author{Weiming Liu,
        Bin Li,{\rm \printfnsymbol{1}}~\IEEEmembership{Member,~IEEE,}
        and~Julian~Togelius,~\IEEEmembership{Member,~IEEE}
        
\thanks{Weiming Liu is with the School of Data Science, University of Science and Technology of China, Anhui, China. (e-mail: weiming@mail.ustc.edu.cn).}
\thanks{Bin Li (corresponding author) is with the School of Information Science and Technology, University of Science and Technology of China, Anhui, China. (e-mail: binli@ustc.edu.cn).}
\thanks{Julian Togelius is with the Department of Computer Science and Engineering, New York University, New York, NY 11201, USA (e-mail: julian@togelius.com).}

\thanks{The work is partially supported by the National Natural
Science Foundation of China under grand No.U19B2044 and
No.61836011.}
}

\markboth{Model-Free Neural CFR with Bootstrap Learning}%
{}

\maketitle

\begin{abstract}
Counterfactual Regret Minimization (CFR) has achieved many fascinating results in solving large-scale Imperfect Information Games (IIGs). Neural network approximation CFR (neural CFR) is one of the promising techniques that can reduce computation and memory consumption by generalizing decision information between similar states. Current neural CFR algorithms have to approximate cumulative regrets. However, efficient and accurate approximation in a large-scale IIG is still a tough challenge. In this paper, a new CFR variant, Recursive CFR (ReCFR), is proposed. In ReCFR, Recursive Substitute Values (RSVs) are learned and used to replace cumulative regrets. It is proven that ReCFR can converge to a Nash equilibrium at a rate of $O({1}/{\sqrt{T}})$. Based on ReCFR, a new model-free neural CFR with bootstrap learning, Neural ReCFR-B, is proposed. Due to the recursive and non-cumulative nature of RSVs, Neural ReCFR-B has lower-variance training targets than other neural CFRs. Experimental results show that Neural ReCFR-B is competitive with the state-of-the-art neural CFR algorithms at a much lower training cost.
\end{abstract}

\begin{IEEEkeywords}
Game Theory, Imperfect Information Games, Counterfactual Regret Minimization, Neural Networks.
\end{IEEEkeywords}

\IEEEpeerreviewmaketitle

\section{Introduction}

\IEEEPARstart{I}{mperfect} Information Game (IIG) is a kind of challenging game, in which players can only obtain partial information. Conventional methods for Perfect Information Games (PIGs) \cite{TDGammon,MCTS,DQN,AlphaGoZero} generally do not play IIGs well \cite{ISMCTS,NestedSubgame,ACIIG}. Usually, IIGs are solved using equilibrium-finding algorithms \cite{CFR,FSP,Oracle,EGT}. This paper focuses on two-player, zero-sum IIGs in extensive form, for example, heads-up no-limit Texas hold’em poker (HUNL). In recent years, Counterfactual Regret Minimization (CFR) algorithm \cite{CFR} has been applied in human-level poker agents \cite{DeepStack,Libratus,Pluribus}. CFR is an iterative algorithm. It has been proven that the average strategy will converge to a Nash equilibrium if the zero-thresholded regret, $\max\{\max_a R^T_p(I, a), 0\}$, at every state increases sub-linearly \cite{CFR}. Here $R^T_p(I, a) = \sum_{t=1}^{T}r^t(I, a)$ is the \textit{cumulative regret}, and $r^t(I, a)$ is the \textit{instantaneous regret}, will be defined later. 
To minimize the cumulative regret at every state, a CFR algorithm needs to choose a strategy at every iteration according to the regrets. Conventional tabular CFR visits all states at every iteration and stores the cumulative regrets in a table. So it is computation expensive and memory intensive in large-scale IIGs.

To deal with large-scale IIGs, the technique ``abstraction'' \cite{eval_abs} was proposed to cluster states and actions. However, abstraction techniques are domain-specific and depend on expert knowledge. Function approximation CFR (FCFR)  \cite{RCFR,morrill2016using} is a kind of algorithm that approximates cumulative regrets with function approximators. Recently, some FCFRs with neural network approximators (neural CFR) have been proposed \cite{DeepCFR,DoubleCFR,dream}. 
Thanks to the potential generalization ability of function approximators, in an FCFR, it can be considered that the states are clustered in an implicit space. In other words, FCFR may abstract a game automatically, and, hopefully, reducing the computation and memory consumption.

Since it is intractable to traverse the full game tree and approximate the cumulative regrets directly in a large-scale IIG, FCFR usually trains the approximator on surrogate training targets. 
As proven in \cite{morrill2016using,DeepCFR}, the quality of the approximation directly impacts the quality of the output strategy. However, the training cost for approximating cumulative regrets could be high.
Take a linear approximator with $p$ parameters as an example. Assume the variance in training targets is $\sigma^2$, and the data set size is $m$. It is known that the variance in approximated values on the fixed data set is $\frac{p}{m}\sigma^2$ \cite{HastieTF09}. Unfortunately, the variance in training targets in a zero-sum IIG is likely to be high due to the adversarial nature.
So, as a compromise, the data set should be large. Also, the approximator should be trained for many epochs to prevent underfitting. 
Therefore, the cost for approximating cumulative regrets could be high.
We will discuss more in Section \ref{sec:back_app}. 

Furthermore, existing neural CFR algorithms usually require an exact simulator for model-based sampling (sampling multiple actions in a single state), in order to reduce the variance in training targets. To extend CFR to more IIGs with unknown models, it is necessary to develop model-free algorithms.

To reduce the training cost and use model-free sampling, it is desired to avoid approximating cumulative regrets.
In this paper, a new CFR algorithm, Recursive CFR (ReCFR), is proposed, in which cumulative regrets are replaced by
non-cumulative Recursive Substitute Values (RSVs) proposed in Warm CFR \cite{StrategyBase}. 
We prove that ReCFR converges to a Nash equilibrium at a rate of $O({1}/{\sqrt{T}})$. We also prove that vanilla CFR and Full-width extensive-form fictitious play (XFP) \cite{FSP} are special cases of ReCFR.

Based on ReCFR, a new neural CFR algorithm, Neural ReCFR with Bootstrapping (Neural ReCFR-B), is proposed. 
In the algorithm, two RSV networks (one for each player) are trained to approximate RSVs. The training targets for RSVs can be estimated by bootstrap learning, which is an important benefit brought by the recursive nature of RSVs (cumulative regrets do not have this property). 
Since RSVs are non-cumulative, the training targets should have lower variance than that for cumulative regrets. Also, bootstrap learning can reduce the variance, at the cost of being a bit biased \cite{RLIntro}. So, due to the lower variance training targets, Neural ReCFR-B may use model-free sampling and train the neural networks with fewer samples at every iteration than other neural CFR algorithms. 
Experimental results show that Neural ReCFR-B is competitive with the state-of-the-art neural CFR algorithms \cite{DeepCFR,DoubleCFR,dream} at a much lower training cost.

The contributions of the paper are mainly in three aspects:
\begin{itemize}
    \item RSVs and Warm CFR \cite{StrategyBase} are revisited, and a new exploitability bound is proven. Two new and intuitive lemmas (Lemma \ref{lm:v_eq_v_r} and Lemma \ref{lm:rec_v_eq_v_r}) are also proposed for analyzing CFR algorithms.
    \item A new CFR algorithm based on RSVs is presented, and it is proven to converge to a Nash equilibrium at a rate of $O({1}/{\sqrt{T}})$. It is also shown that ReCFR can generalize both Fictitious Play \cite{FSP} and vanilla CFR.
    \item A new model-free neural CFR algorithm based on ReCFR is presented, and it shows higher training efficiency than other algorithms on medium-size and large-size IIGs.
\end{itemize}

In the rest of the paper, related work is first discussed in Section \ref{sec:ret}. Then, notations and background are given in Section \ref{sec:notation}. 
In Section \ref{sec:revisit}, RSVs and Warm CFR are revisited, and two new lemmas are given.
In Section \ref{sec:algorithm} and \ref{sec:ncfrb}, ReCFR and Neural ReCFR-B are described in detail. Then, the experimental setup and the results of ReCFR and Neural ReCFR-B are shown in Section \ref{sec:setup}. Finally, the conclusion is drawn in Section \ref{sec:conclusion}.

\section{Related Work}
\label{sec:ret}
In recent years, many techniques have been proposed to improve the performance of CFR in large-scale IIGs \cite{CFR-D,NestedSubgame,Depthlimited,DeepStack}.
Based on these techniques, the state-of-the-art algorithms for HUNL are Libratus \cite{Libratus} and Pluribus \cite{Pluribus}. 

Regression CFR \cite{RCFR} is the first FCFR.
Deep CFR \cite{DeepCFR} and Double Neural CFR (DNCFR) \cite{DoubleCFR} are two more recent neural CFRs. 
Deep CFR uses two reservoir memory buffers to store all past instantaneous regrets and uses two regret networks (one for each player) to approximate the mean values of the past regrets. However, due to the high variance training targets, the variance in approximated cumulative regrets could be high. 
Another neural CFR, DNCFR, sums the current-iterate instantaneous regrets and the previous approximation as the training targets. However, the approximation error in each iteration will accumulate. So, as the number of iterations goes to infinity, the bias is unbounded. The problem may be alleviated using Regret Matching+ algorithm \cite{CFRPlus} and model-based robust sampling \cite{DoubleCFR}, but there is no theoretical guarantee. 
Single Deep CFR \cite{SingleDeepCFR} is a variant of Deep CFR, which does not need to approximate the average strategy. Instead, it keeps all regret networks of past iterations and computes an average strategy at the end. However, cumulative regrets are still approximated.

Neural Fictitious Self-Play (NFSP) \cite{NFSP}  is a model-free algorithm  based on full-width extensive-form fictitious play (XFP) \cite{FSP}. 
NFSP uses DQN \cite{DQN} to learn the best-response strategy for each player at each iteration and uses a strategy network to approximate the average best-response strategy.
Since CFR has a better convergence guarantee than XFP, 
people may also be interested in model-free neural CFR algorithms.
ARMAC \cite{ARMAC} and DREAM \cite{dream} are both model-free neural CFR algorithms like Neural ReCFR-B. However, both of them approximate cumulative regrets. Specifically, ARMAC approximates cumulative regrets by replaying past strategies and regenerating past regrets, while DREAM uses a reservoir buffer to store all past regrets.  Besides, they also train an additional state-action network to reduce the variance in training targets \cite{VRMCCFR}. 
A critical difference of Neural ReCFR-B from these two methods is that it does not store or regenerate past regrets but learns RSVs only based on the average strategy.

Some other works have combined equilibrium-finding algorithms with gradient descending \cite{ACPO,ED}. They have achieved performance comparable to NFSP, and they seem to have the potential for solving large-scale IIGs. However, they have only been tested in small-scale games.

\section{Notations and Background}
\label{sec:notation}
For a two-player zero-sum IIG, the set of players is denoted by $P=\{1,2\}$. The chance player, denoted by $c$, is introduced to take actions for random events.
A history $h$ is represented as a sequence of actions taken by all the players and the chance player. The root history is represented as an empty sequence, written as $\emptyset$. Denote the set of histories by $H$. For any non-terminal history $h\in H$, the set of legal actions at $h$ is denoted by $A(h)$. The set of all actions is denoted by $\mathcal{A}$.  The acting player at history $h$ is denoted by $P(h)$, where $P(h)\in P\cup \{c\}$. 
In this paper, we assume that $P(\emptyset) \in P$ and will present the main results under the viewpoint of $p = P(\emptyset)$.\footnote{For the other player $p' \neq P(\emptyset)$ we can construct an equivalent surrogate game with $P(\emptyset) = p'$ by adding a player $p'$ decision point at the beginning of the game with a single action \cite{StrategyBase}.}
After player $P(h)$ takes an action $a\in A(h)$, the resulted history is denoted by $h \cdotp a$. If there exists a sequence of actions from $h$ to $h'$, then $h'$ is a descendent of $h$, denoted by $h\sqsubset h'$. Let $h \sqsubseteq h'$ represent that $h \sqsubset h'$ or $h = h'$. 
We only consider depth-limited games in this paper. The set of terminal histories is denoted by $Z$. For any $z\in Z$, the \textit{payoff} for player $p$ is $u_p(z)\in \mathbb{R}$. As the game is two-player and zero-sum, we have $u_1(z)=-u_2(z)$. 
Define $ Succ_p(h)$ as the  set of the earliest reachable histories such that for any $h'\in Succ_p(h)$, $P(h') = p$ or $h' \in Z$. ``Earliest reachable'' means that $h \sqsubseteq h'$ and there is no $h''$ such that $h''\sqsubset h'$.

In an imperfect information game, the histories of each player $p \in P$ are divided into \textit{information sets} (infosets). The set of infosets of player $p$ is denoted by $\mathcal{I}_p$. 
Let $\mathcal{I} = \mathcal{I}_1 \cup \mathcal{I}_2$. 
For any infoset $I \in \mathcal{I}_p$, all histories $h, h'\in I$ are indistinguishable to $p$, so $A(h) = A(h')$. 
Define $P(I)=P(h)$ and $A(I)=A(h)$ for any $h\in I$. Let $I(h)$ denote the infoset of history $h$, i.e., $I(h) = I$ for any $h\in I$.
Besides, define $Succ_p(I,a)$ as the set of the earliest reachable infosets after action $a$ has been taken. Formally, $Succ_p(I, a) = \{I(h') | h' \notin Z,h'\in Succ_p(h \cdotp a), h\in I\}$. Let $Succ_p(I) = \bigcup_{a\in A(I)} Succ_p(I, a)$.
The range of payoffs reachable from $I$ is denoted by $\D(I)$. Formally, $\D(I) = \max_{z\in Z, h\in I:h\sqsubset z}u_{P(I)}(z) - \min_{z\in Z, h\in I:h\sqsubset z}u_{P(I)}(z)$. Let $\D = \max_{I\in \mathcal{I}} \D(I)$.

A strategy $\sigma_p$ of a player $p \in P$ is a function that maps any infoset $I\in \mathcal{I}_p$ to a probability vector over $A(I)$. The set of strategies of player $p$ is denoted by $\Sigma_p$. Given a strategy $\sigma_p\in \Sigma_p$, $\sigma_p(I)$ is the probability vector, and $\sigma_p(I, a)$ is the probability of choosing action $a$ at infoset $I$. Since the histories in an infoset are indistinguishable, the strategy in each of them must be identical. So $\sigma_p(h) = \sigma_p(I)$ and $\sigma_p(h, a) = \sigma_p(I, a)$ for any $h\in I$ and $a\in A(I)$. The strategy of the other player (the opponent) is denoted by $\sigma_{-p}$. Denote a strategy profile by $\sigma=\langle\sigma_p, \sigma_{-p}\rangle$.  The chance player's strategy, which is fixed and known to all the players, is denoted by $\sigma_c(h, a)$.

$\pi^\sigma(h)=\prod_{h' a \sqsubseteq h}{\sigma_{P(h')}(h', a)}$ is called the \textit{reach} of $h$, which is the probability of reaching $h$ when all the players act according to $\sigma$.  $\pi_{p}^\sigma(h)$ is the contribution of $p$ to this probability. $\pi_{-p}^\sigma(h)$ is the contribution of the opponent and the chance player. The probability of reaching $h'$ from $h$ is denoted by $\pi^\sigma(h, h')$. 
In this paper, we only consider \textit{perfect recall} games. Therefore, we
define $\pi^\sigma_p(I) = \pi^\sigma_p(h)$ for any $h\in I$. Accordingly, define the reach of the opponent and the chance player as $\pi^\sigma_{-p}(I) = \sum_{h\in I}\pi^\sigma_p(h)$. 
The \textit{expected payoff} of the game for player $p$ is denoted by $u_p(\sigma_p, \sigma_{-p})$. Formally, $u_p(\sigma_p, \sigma_{-p}) = \sum_{z\in Z}{\pi^\sigma(z)u_p(z)}$.
A \textit{best response} $\BR(\sigma_{-p})$ is a strategy of player $p$ such that $\BR(\sigma_{-p}) = \argmax_{\sigma'_p \in \Sigma_p}{u_p(\sigma'_p, \sigma_{-p})}$. 
A \textit{Nash equilibrium} $\sigma^*=\langle \sigma_p^*, \sigma_{-p}^* \rangle$ is a strategy profile where every player plays a best response. The \textit{exploitability} of a strategy $\sigma_p$ is the distance to a Nash equilibrium, defined as $e(\sigma_p) = u_{p}(\sigma_p^*, \BR(\sigma_p^*)) - u_{p}(\sigma_p, \BR(\sigma_p))$. Define the total exploitability as $e(\sigma) = \sum_{p\in P}{e(\sigma_p})$. 

\subsection{Counterfactual Regret Minimization (CFR)}

CFR is an iterative algorithm for two-player zero-sum IIGs. It computes a strategy profile at every iteration using Regret Matching (RM) algorithm \cite{CFR} according to the cumulative regrets of infosets. Then, the full game tree is traversed, and the cumulative regrets and average strategy are updated according to the strategy. It has been proven that the total exploitability of the average strategy is bounded by $O({1}/{\sqrt{T}})$ after $T$ iterations of CFR \cite{CFR} are played.

Let $\sigma^t$ be the strategy at iteration $t$.
The \textit{counterfactual value} of action $a$ at infoset $I$ is defined as
\begin{equation}
\label{eq:cfr_v}
v_p^{\sigma^t}(I, a) = \sum_{h\in I}\sum_{z\in Z:h\sqsubset z}\pi^{\sigma^t}_{-p}(h)\pi^{\sigma^t}(h \cdotp a, z)u_p(z).
\end{equation}
$v^{\sigma^t}_p(I) = \sum_{a\in A(I)}\sigma^t_p(I, a)v_p^{\sigma^t}(I, a)$ is the counterfactual value of infoset $I$. Counterfactual values can also be defined recursively: at each infoset $I$,
\begin{equation}
\label{eq:rec_v}
\begin{aligned}
v^{\sigma^t}_p(I, a) = &\sum_{h\in I}{\sum_{z\in Z: z \in Succ_p(h \cdotp a)}{\pi_{-p}^{\sigma^t}(z)u_p(z)}} + \\
& \sum_{I'\in Succ_p(I, a)}{v^{\sigma^t}_p(I')}.
\end{aligned}
\end{equation}
This equation has been used in some existing literature \cite{CFR,StrategyBase}. A proof is also provided in Appendix A.\footnote{Appendix: https://arxiv.org/abs/2012.01870}
Let $v^{\sigma^t}_p =v^{\sigma^t}_p(I(\emptyset))$, then, $u_p(\sigma^t_p, \sigma^t_{-p}) = v^{\sigma^t}_p$.
In the rest of the paper, we mainly use $v^{\sigma^t}_p$ and $v^{\langle\sigma^t_p, \sigma^t_{-p}\rangle}_p$ to denote the expected payoff.
The \textit{instantaneous regret} is defined as $r_p^t(I, a) = v_p^{\sigma^t}(I, a) - v_p^{\sigma^t}(I)$. 
The \textit{cumulative counterfactual regret} (\textit{cumulative regret}) of action $a$ at $I$ is
\begin{equation}
\label{eq:cum_cr}
R^T_p(I, a) = \sum_{t=1}^T{v_p^{\sigma^t}(I, a)} - \sum_{t=1}^T{v_p^{\sigma^t}(I)}.
\end{equation}
The \textit{average strategy} $\overline{\sigma}^T_p(I, a)$ is computed\footnote{We assume $\overline{\sigma}^t(I, a) > 0$ (and thus $\pi^{\overline{\sigma}^t}(I) > 0$) for any $I \in \mathcal{I}$ and  $t \geq 1$. This is true if, e.g., $\sigma^1(I, a) = \frac{1}{|A(I)|} > 0$ for every $I\in \mathcal{I}$.} according to
\begin{equation}
\overline{\sigma}^T_p(I, a) = \frac{\sum_{t = 1}^{T}{\pi_p^{\sigma^t}(I)\sigma^t_p(I, a)}}{\sum_{t = 1}^{T}{\pi_p^{\sigma^t}(I)}}.
\label{eq:ave_strategy}
\end{equation}

The \textit{total regret} of player $p$ after $T$ iterations is 
\begin{equation}
\label{eq:total_regret}
R_p^T = \max_{\sigma_p' \in \Sigma_p}{\sum_{t=1}^T{v^{\langle\sigma_p', \sigma_{-p}^t\rangle}_p} - \sum_{t=1}^T{v^{\sigma^t}_p}}.
\end{equation} 
It has been proven in \cite{CFR} that $R^T_p \leq \sum_{I\in \mathcal{I}_p} \max_{a}(R^T_p(I, a))_+$, 
where $(\cdot)_+ = \max\{\cdot, 0\}$. Therefore, the total regret $R^T_p$ can be minimized by minimizing $\max_{a}(R^T_p(I, a))_+$ at every infoset, using, e.g., RM. 
At each infoset, RM computes the next iteration strategy according to
\begin{equation}
\sigma^{t + 1}_p(I, a) = 
\frac{(R^t_p(I, a))_+}{\sum_{a' \in A(I)}{(R^t_p(I, a'))_+}}.
\label{eq:rm}
\end{equation}
If \(\sum_{a'}{(R^t_p(I, a'))_+} = 0\), the action with the highest cumulative regret is assigned with probability 1 \cite{DeepCFR}. According to \cite{StrategyBase}, we have the following lemma. 
\begin{lemma}
\label{lm:sb_R_le_L}
\cite{StrategyBase}
After T iterations of CFR are played, for any infoset $I$,
$
\sum_{a\in A(I)}{\left(R^T_p\left(I, a\right)\right)_+^2 \leq \pi_{-p}^{\overline{\sigma}^T}(I) \D^2(I)|A\left(I\right)|T}.
$
\end{lemma}
Finally, it is well known that, in a two-player zero-sum game, the total exploitability equals the average total regret, i.e,. $\epsilon(\overline{\sigma}^T) =  \frac{1}{T}\sum_{p\in P}R^T_p$. This can be seen by the definitions. Therefore, $\epsilon(\overline{\sigma}^T) = O({1}/{\sqrt{T}})$ in CFR \cite{CFR}.

\subsection{ Recursive Substitute Values (RSVs) and Warm CFR}

RSVs were first proposed in Warm CFR \cite{StrategyBase}. 
Given an arbitrary strategy $\sigma$, Warm CFR initializes the cumulative regrets with substitute regrets, hoping the CFR algorithm can converge faster.
The initialization is done by treating the strategy $\sigma$ as the average strategy after $T$ iterations of CFR are run. To some extent, the initialization is equivalent to recovering the cumulative regrets. Let $v'^{\sigma}_p(I, a)$ and $v'^{\sigma}_p(I)$ be the RSVs that recover the true average counterfactual values:  $\frac{1}{T}\sum_{t=1}^T{v_p^{\sigma^t}(I, a)}$ and $ \frac{1}{T}\sum_{t=1}^T{v_p^{\sigma^t}(I)}$. Then, the \textit{substitute cumulative counterfactual regret} (\textit{substitute regret}), 
\begin{equation}
\label{eq:sb_R}
R'^T_p(I, a) = T \left(v'^{\sigma}_p\left(I, a\right) - v'^{\sigma}_p\left(I\right)\right),
\end{equation}
should recover $R^T_p(I, a)$ at every infoset. 
According to the definition of counterfactual values and Lemma \ref{lm:sb_R_le_L}, the RSVs should satisfy the two constraints below if they are the true average counterfactual values:
\begin{equation}
\label{eq:rec_sub_v}
\begin{aligned}
    v'^{\sigma}_p(I, a) = &\sum_{h\in I}{\sum_{z\in Z: z \in Succ_p(h \cdotp a)}{\pi_{-p}^{\sigma}(z)u_p(z)}} + \\
    &\sum_{I'\in  Succ_p(I, a)}{v'^{\sigma}_p(I')},
\end{aligned}
\end{equation}
\begin{equation}
\label{eq:sb_cons}
\sum_{a\in A(I)}{\left(v'^{\sigma}_p\left(I, a\right) -  v'^{\sigma}_p\left(I\right)\right)_+^2}\leq  \frac{\pi^{\sigma}_{-p}(I)\D^2(I) \left|{A(I)}\right|}{T}.
\end{equation}
There could be many RSV profiles that fulfill the two constraints above. However, we can choose the RSVs for each infoset recursively: at infoset $I$, $v'^{\sigma}_p(I, a)$ is computed according to the immediate payoffs and the RSVs of its earliest reachable infosets, while $v'^{\sigma}_p(I)$ is chosen to fulfill (\ref{eq:sb_cons}).\footnote{Because of the non-smooth zero-thresholding operator, $v'^{\sigma}_p(I)$ can not be presented in an explicit form. More details are provided in Appendix A.}
Define the \textit{substitute expected payoff} as
\begin{equation}
    v'^{\sigma}_p = v'^{\sigma}_p(I(\emptyset)).
\end{equation}
In Warm CFR, another $T'$ CFR iterations are run based on the substitute regrets. Specifically, define
\begin{equation}
\label{eq:sub_regret}
    R'^{T, T'}_p(I, a) = R'^T_p(I, a) + \sum_{t'=1}^{T'}r^{\sigma^{t'}}_p(I, a).
\end{equation}
The strategy $\sigma^{t'}_p$ at iteration $t'$ is computed using RM according to $R'^{T, t' - 1}_p(I, a)$.
Define the average strategy as
$
    \overline{\sigma}^{T,T'}_p = \frac{1}{T + T'}{(T \sigma_p +  T' \overline{\sigma}^{T'}_p)},
$
where $\overline{\sigma}^{T'}_p$ is the average strategy of the $T'$ iterations.
It is proven in \cite{StrategyBase} that 
\begin{equation}
\label{eq:sb_R_bound}
    \sum_{a\in A(I)}{\left(R'^{T, T'}_p(I, a)\right)_+^2 \leq \pi^{\overline{\sigma}^{T,T'}}_{-p}(I) \D^2(I)|A\left(I\right)|(T+T')}.
\end{equation}
As a result, the average strategy will converge to a Nash equilibrium at a rate of $O({1}/{\sqrt{T + T'}})$ under specific condition (Assumption \ref{ass:2}, will be discussed in the next section). 

\subsection{Monte Carlo CFR}
Monte Carlo CFR (MCCFR) \cite{MCCFR} is a variant of CFR that only traverses parts of the game tree at every iteration and estimates the instantaneous regrets by a Monte Carlo (MC) method. At iteration $t$, suppose a subset $Q$ of the full game tree is sampled, the sampled counterfactual value at $(I, a)$ is defined as
\begin{equation}
\tilde{v}^{\sigma^t}_p(I, a) = \sum_{z\in Q \cap Z}{ \frac{1}{q(z)}\pi_{-p}^{\sigma^t}(h)\pi^{\sigma^t}(h \cdotp a, z)u_p(z)},
\end{equation}
where $q(z)$ is the probability of sampling $z$.
Define $\tilde{v}^{\sigma^t}_p(I) = \sum_{a\in A(I)}\sigma^t_p(I, a)\tilde{v}_p^{\sigma^t}(I, a)$.
The sampled instantaneous regret is $\tilde{r}^{\sigma^t}_p(I, a) = \tilde{v}^{\sigma^t}_p(I, a) - \tilde{v}^{\sigma^t}_p(I)$. The sampled cumulative regret is $\tilde{R}^T_p(I, a) = \sum_{t=1}^T{\tilde{r}^{\sigma^t}_p(I, a)}$, which is an unbiased estimator of the cumulative regret.
There are many MCCFR algorithms \cite{MCCFR,publicMCCFR,generalMCCFR}, of which External Sampling CFR (ESCFR) and Outcome Sampling CFR (OSCFR) proposed in \cite{MCCFR} are the most common. 

\subsection{Function Approximation of Cumulative Regrets}
\label{sec:back_app}
For a two-player IIG, FCFR is usually combined with MCCFR and approximates the MC sampled cumulative regrets by training on surrogate training targets \cite{DeepCFR,DoubleCFR,dream}.
Denote the training target at $(I, a)$ by $\hat{R}^T(I, a)$ and assume $\hat{R}^T(I, a) = \tilde{R}^T(I, a) + \epsilon$, where $\tilde{R}^T(I, a)$ is the MC sampled cumulative regret and $\epsilon$ is a random variable at the infoset.
The variance in training targets is defined as $\operatorname{var}{(\hat{R}^T)} = \mathbb{E}_{I, a, \epsilon}(\hat{R}^T(I, a) - \mathbb{E}_{\epsilon}\hat{R}^T(I, a))^2$.
The expected error is  ${Err} = \mathbb{E}_{I, a, D}(f^T(I, a|D) - \tilde{R}^T(I, a))^2$.
According to the famous bias-variance decomposition, 
$
Err = (\operatorname{bias}(f^T))^2 + \operatorname{var}(f^T)
$,
where $(\operatorname{bias}(f^T))^2$ and $\operatorname{var}(f^T)$ are the bias and variance in approximated regrets, respectively. Specifically, $(\operatorname{bias}(f^T))^2 = \mathbb{E}_{I, a}(\mathbb{E}_D f^T(I, a|D) - \tilde{R}^T(I, a))^2$ and  $\operatorname{var}(f^T) = \mathbb{E}_{I, a, D}(f^T(I, a|D) - \mathbb{E}_D f^T(I, a|D))^2$. 
The bias could be low if the training targets are unbiased estimators. However, as we mentioned before, the variance in approximated cumulative regrets is affected by the variance in training targets and data set size.

As discussed in existing literature \cite{morrill2016using,DeepCFR,DoubleCFR}, there are at least two methods with different training targets for approximating sampled cumulative regrets. Note that a regret $\tilde{r}^{\sigma^t}(I, a), t = 1,\dots, T$ sampled according to MCCFR is an unbiased estimator of a kind of average regret at infoset $I$. So, the first method, which has been used in Deep CFR and DREAM, is to train on a data set $D = \{(I, \tilde{r}^{\sigma^t}(I, \cdot)) |x^t(I) = 1, t=1,\dots, T\}$, where $x^t(I)$ is an indicator variable that is 
$1$ if and only if $I$ is sampled at iteration $t$, and it is equal to $0$ otherwise.
Due to the adversarial nature of zero-sum IIGs, the sampled instantaneous regrets from different iterations are non-IID, and their variance tends to be high. To reduce the variance in training targets, model-based sampling is used in Deep CFR, and a variance reduction technique \cite{VRMCCFR} is used in DREAM. A large regret memory buffer is also used.
Besides, considering that the one-to-many mapping $(I, a) \mapsto \tilde{r}^{\sigma^t}(I, a)$ is complex, they train the approximators for many epochs to prevent underfitting.

The second method,  which was first proposed in \cite{morrill2016using} and has been used in DNCFR \cite{DoubleCFR}, is to estimate $\tilde{R}^{T}(I, a)$ at iteration $T$ with $f^{T-1}(I, a) + \tilde{r}^{\sigma^T}(I, a)$ and train the approximator on a data set $D = \{(I, f^{T-1}(I, \cdot) + \tilde{r}^{\sigma^T}(I, \cdot))|I\in x^t(I)\}$.
So, this is a method bootstrapping on the last approximation. The variance in training targets should be low since variance only comes from the current-iterate instantaneous regrets.
However, the targets are \textit{biased} estimators of the sampled cumulative regrets, and the approximation error can \textit{accumulate} over iterations. In other words, the bias in approximated cumulative regrets is unbounded as the number of iterations goes to infinity. As a compromise, DNCFR trains the approximator for many epochs at every iteration.
In conclusion, the cost of approximating cumulative regrets is high.

\section{Revising RSVs and Warm CFR}
\label{sec:revisit}
First, we would like to present a new and intuitive lemma for analyzing CFR algorithms. 
\begin{lemma}
\label{lm:v_eq_v_r}
For any strategy profile $\sigma = \langle\sigma_p, \sigma_{-p}\rangle$, and another strategy  $\sigma'_p$ of player $p$, we have
$
v^{\langle\sigma'_p, \sigma_{-p}\rangle}_p - v^\sigma_p  
=  \sum_{I\in \mathcal{I}_p}\sum_{a\in A(I)}{\pi^{\sigma'}_p(I)\left(v^{\sigma}_p(I, a) - v^{\sigma}_p(I) \right) \sigma'_p(I, a)}.
$
\end{lemma}

The lemma utilizes the recursive definition of counterfactual values (Equation (\ref{eq:rec_v})). 
Lemma \ref{lm:v_eq_v_r} shows that the difference between expected payoffs can be completely represented by instantaneous regrets. As an application, the famous inequality $R^T_p \leq  \sum_{I\in \mathcal{I}_p} \max_{a}(R^T_p(I, a))_+$ \cite{CFR} is immediately recovered because 
${\sum_{t=1}^T{v^{\langle\sigma_p', \sigma_{-p}^t\rangle}_p} - \sum_{t=1}^T{v^{\sigma^t}_p}} 
= {\sum_{I\in \mathcal{I}_p}\sum_{a\in A(I)}{\pi^{\sigma'}_p(I)\left(\sum_{t=1}^Tr^t_p(I, a)\right) \sigma'_p(I, a)}}$.

Since RSVs have Equation (\ref{eq:rec_sub_v}) similar to (\ref{eq:rec_v}), we have Lemma \ref{lm:rec_v_eq_v_r}. 
Both lemmas are proven in Appendix A.
\begin{lemma}
\label{lm:rec_v_eq_v_r}
For any strategy $\sigma = \langle\sigma_p, \sigma_{-p}\rangle$, another strategy $\sigma'_p $ of player $p$, and arbitrary $v'^\sigma_p(I)$ at all infosets, compute $v'^\sigma_p(I, a)$ according to (\ref{eq:rec_sub_v}), then, 
$
 v^{\langle\sigma'_p, \sigma_{-p}\rangle} - v'^{\sigma}_p 
=  \sum_{I\in \mathcal{I}_p}\sum_{a\in A(I)}{\pi^{\sigma'}_p(I)\left(v'^{\sigma}_p(I, a) - v'^{\sigma}_p(I) \right) \sigma'_p(I, a)}.
$
\end{lemma}
 
Note that Lemma \ref{lm:rec_v_eq_v_r} does not rely on how $v'^\sigma_p(I)$ is chosen at each infoset.
Now, let us go back to the setting of Warm CFR. Assume the initial strategy $\sigma$ is an average strategy $\overline{\sigma}^T$ generated by an arbitrary iterative algorithm, and another $T'$ iterations of CFR are run based on the substitute regrets. As a result of Lemma \ref{lm:v_eq_v_r} and \ref{lm:rec_v_eq_v_r}, the total regret of the $T + T'$ iterations is
\begin{equation}
\label{eq:r_decomp}
\begin{aligned}
& R^{T + T'}_p =  \max_{\sigma_p' \in \Sigma_p}{\sum_{t=1}^{T+T'}{v^{\langle\sigma_p', \sigma_{-p}^t\rangle}_p} - \sum_{t=1}^{T+T'}{v^{\sigma^t}_p}} \\
\leq & \underbrace{\left(T v'^{\sigma}_p - \sum_{t=1}^T{v^{\sigma^t}_p}\right)}_{\textbf{Term}_1} + \underbrace{\sum_{I\in \mathcal{I}_p}\max_a{(R'^{T,T'}_p(I, a))_+}}_{\textbf{Term}_2}.
\end{aligned}
\end{equation}
A proof is provided in Appendix A. 
In the equation, the total regret of the $T + T'$ iterations is \textit{decomposed} to two terms related to the RSVs. 
Specifically,  $\textbf{Term}_1$ measures the difference between the substitute expected payoff and the true expected payoffs in the first $T$ iterations, while $\textbf{Term}_2$ is the sum of the substitute regrets. 

\begin{assumption}
\label{ass:2}
\cite{StrategyBase}
$
    v'^{\sigma}_1 + v'^{\sigma}_2 \leq 0
$.
\end{assumption}

As shown in \cite{StrategyBase}, if Assumption \ref{ass:2} is true, the total regret will be bounded by the sum  of the substitute regrets, i.e., $\sum_{p\in P}R^{T + T'}_p \leq \sum_{p\in P} \sum_{I\in \mathcal{I}_p} \max_a (R'^{T,T'}_p(I, a))_+$.
Note that $\sum_{t=1}^T{v^{\sigma^t}_p}$ in $\textbf{Term}_1$ in (\ref{eq:r_decomp}) is canceled out as $v^{\sigma^t}_1 + v^{\sigma^t}_2 = 0$.
Consequently, according to (\ref{eq:sb_R_bound}), we have $\epsilon(\overline{\sigma}^{T, T'}) = \frac{1}{T + T'}\sum_{p\in P}R^{T + T'}_p =  O({1}/{\sqrt{T+T'}})$. 
However, is Assumption  \ref{ass:2} necessary for convergence? The answer may be no. Note that $\textbf{Term}_1$ in (\ref{eq:r_decomp}) does not depend on $T'$, and it may be bounded by the total regret of the initial $T$ iterations. Based on these observations, we have the following theorem.

\begin{theorem}
\label{th:new_sub_bound}
Given an arbitrary  strategy $\sigma$ and $T > 0$, choose the RSVs according to (\ref{eq:rec_sub_v}) and (\ref{eq:sb_cons}) and assume $v'^{\sigma}_p\left(I\right) \leq \max_{a}v'^{\sigma}_p\left(I, a\right)$ at every infoset. If another $T'$ iterations of CFR are run based on the substitute regrets, then, 
$
\epsilon(\overline{\sigma}^{T,T'}) \leq \frac{T\epsilon(\sigma)}{T + T'} + \frac{1}{\sqrt{T + T'}}\sum_{I\in \mathcal{I}_p}{\sqrt{\pi^{\overline{\sigma}^{T,T'}}_{-p}(I)} \D(I)\sqrt{|A(I)|}}.
$
\end{theorem}

The proof is given in Appendix A.
As we can see, when $T'$ is sufficiently large (e.g., $T' \geq (T\epsilon(\sigma))^2 - T$), the first term $T\epsilon(\sigma)/({T + T'}) = O({1}/{\sqrt{T + T'}})$, and thus $\epsilon(\overline{\sigma}^{T,T'}) = O({1}/{\sqrt{T + T'}})$. 
For example, if the initial strategy is an average strategy of a CFR algorithm, we have $\epsilon(\sigma) = O({1}/{\sqrt{T}})$ and $e(\sigma^{T,T'}) = O(\sqrt{T}/({T + T'}) + {1}/{\sqrt{T+T'}}) = O({1}/{\sqrt{T + T'}})$ for any $T' > 0$. 
Note that Theorem \ref{th:new_sub_bound} does not rely on Assumption \ref{ass:2}. 

In this section, we relax Assumption \ref{ass:2} in Warm CFR to a trivial constraint: $v'^{\sigma}_p\left(I\right) \leq \max_{a}v'^{\sigma}_p\left(I, a\right)$ for all infosets.
Following the idea, we can prove the convergence of ReCFR given in the next section.

\section{A New CFR Algorithm}
\label{sec:algorithm}
In this section,  a new CFR algorithm, named Recursive CFR (ReCFR), is presented. 
ReCFR is based on RSVs proposed in \cite{StrategyBase}.
Instead of only computing the RSVs at the beginning for warm starting, ReCFR discards the cumulative regrets and replaces them with the substitute regrets at every iteration. Therefore, the cumulative regrets are never tracked. 
The pseudocode of ReCFR is given in Algorithm \ref{alg:CFR_sub}.

\begin{algorithm}[htbp]
	\caption{Recursive CFR}
	\label{alg:CFR_sub}
	\begin{algorithmic}[1]
    	\State{Input: maximum iterations $T$.}
    		\For {iteration $t= 1$ to $T$}
    		    \For {player $p \in P$}
                	\For {infoset $I\in \mathcal{I}_p$}
                	    \State $\sigma^{t}_p(I, a) \gets
                	            \begin{cases}
                	                \operatorname{RM}(R'^{t-1}_p(I, a)), &\quad t > 1,\\
                	                \frac{1}{|A(I)|}, &\quad t = 1.
                	            \end{cases}$
            	        \State Update $\overline{\sigma}^t_p(I, a)$.
                	    \Comment{(\ref{eq:ave_strategy})}
                	    \State Compute $v'^{\overline{\sigma}^t}_p(I, a)$ and $v'^{\overline{\sigma}^t}_p(I)$. \Comment{(\ref{eq:rec_sub_v}), (\ref{eq:re_cons})}
                	    \State $R'^t_p(I, a) \gets t(v'^{\overline{\sigma}^t}_p(I, a) - v'^{\overline{\sigma}^t}_p(I))$.
                     \EndFor
                \EndFor
             \EndFor
	\end{algorithmic}
\end{algorithm}

Specifically, at iteration $t$, 
the RSV $v'^{\overline{\sigma}^t}_p\left(I, a\right)$ is computed according to (\ref{eq:rec_sub_v}). However, we choose  $v'^{\overline{\sigma}^t}_p\left(I\right)$ at each infoset according to the constraint
\begin{equation}
\label{eq:re_cons}
\begin{aligned}
\sum_{a\in A(I)}{\left(t v'^{\overline{\sigma}^t}_p\left(I, a\right) - t v'^{\overline{\sigma}^t}_p\left(I\right)\right)_+^2} = \lambda^t_p(I),
\end{aligned}
\end{equation}
where $\lambda^t_p(I) \in [0, \infty) $ is a parameter, will be set later.
Note that when $\lambda^t_p(I) > 0$, the $v'^{\overline{\sigma}^t}_p\left(I\right)$ fulfilling the constraint exists and is unique. Moreover, $v'^{\overline{\sigma}^t}_p\left(I\right) < \max_{a}v'^{\overline{\sigma}^t}_p\left(I, a\right)$. 
When $\lambda^t_p(I)  = 0$, we force that $v'^{\overline{\sigma}^t}_p(I) = \max_{a} v'^{\overline{\sigma}^t}_p(I,a)$.
The substitute regret $R'^{t}_p(I, a)$ is 
\begin{equation}
    R'^t_p(I, a) = t (v'^{\overline{\sigma}^t}_p(I, a) - v'^{\overline{\sigma}^t}_p(I)).
\end{equation}
The strategy at iteration $t + 1$ is computed using RM:
\begin{equation}
\label{eq:re_st}
    \sigma^{t + 1}_p(I, a) = 
\frac{(R'^{t}_p(I, a))_+}{\sum_{a' \in A(I)}{(R'^{t}_p(I, a'))_+}}.
\end{equation}
When $\sum_{a' \in A(I)}{(R'^{t}_p(I, a'))_+} = 0$, the action with the maximal $R'^{t}_p(I, a)$ is assigned with probability 1.
We set $\sigma^1_p(I, a) = {1}/{|A(I)|}$. The average strategy is computed according to (\ref{eq:ave_strategy}).
Since solving (\ref{eq:re_cons}) requires a linear search, the cost of ReCFR at every iteration is $O(|\mathcal{I}||\mathcal{A}|^2)$, which is worse than vanilla CFR ($O(|\mathcal{I}||\mathcal{A}|)$).  

Note that when $\lambda^t_p(I) = 0$ at all infosets, we have $v^{\langle\sigma^{t+1}_p, \overline{\sigma}^t_{-p}\rangle}_p = \max_{\sigma'_p}v^{\langle\sigma'_p, \overline{\sigma}^t_{-p}\rangle}_p$. In other words, $\sigma^{t+1}_p$ is a best response to $ \overline{\sigma}^t_{-p}$.
According to the definition of XFP \cite{FSP}, we have proposition \ref{prop:re_eq_XFP}. 
Similarly, when $\lambda^t_p(I) = \sum_{a}(R^t(I, a))^2_+$ at each infoset, vanilla CFR is recovered. The proofs for the propositions are given in Appendix B.
\begin{proposition}
\label{prop:re_eq_XFP}
ReCFR is equivalent to XFP if $\lambda^t_p(I) = 0$ at every infoset.
\end{proposition}

\begin{proposition}
\label{prop:re_eq_CFR}
ReCFR is equivalent to vanilla CFR if $\lambda^t_p(I) = \sum_{a}(R^t(I, a))^2_+$ at each infoset.
\end{proposition}

\subsection{Properties of Recursive CFR}
According to (\ref{eq:r_decomp}), 
the total regret of ReCFR at iteration $T$ can also be decomposed as 
\begin{equation}
\label{eq:rr_decomp}
R^{T}_p \leq  \underbrace{\left(T v'^{\overline{\sigma}^T}_p - \sum_{t=1}^T{v^{\sigma^t}_p}\right)}_{\textbf{Term}_1} + \underbrace{\sum_{I\in \mathcal{I}_p} \max_a (R'^{T}_p(I, a))_+}_{\textbf{Term}_2}.
\end{equation}
Therefore, if Assumption \ref{ass:2} is true at iteration $T$, $\textbf{Term}_1$ can be canceled out when summing over $R^T_p$ of the players, and thus $\epsilon(\overline{\sigma}^T) \leq \frac{1}{T}\sum_{p\in P}\sum_{I\in \mathcal{I}_p}\sqrt{\lambda^T_p(I)}$ (remember that  $\epsilon(\overline{\sigma}^T) = \frac{1}{T}\sum_{p\in P}R^T_p$).
However, it is \textit{non-trivial} to ensure both Assumption \ref{ass:2} and  ${\lambda^T_p(I)} = O(T)$ \textit{as $T\to \infty$}.\footnote{Corollary 1 in \cite{StrategyBase} does not apply. It assumes $T$ iterations of CFR were played (so Lemma \ref{lm:sb_R_le_L} holds) when computing the RSVs, which is not true in ReCFR.} In other words, we may not be able to guarantee that $\epsilon(\overline{\sigma}^T) = O({1}/{\sqrt{T}})$.
Fortunately, similar to Theorem \ref{th:new_sub_bound}, Assumption \ref{ass:2} is unnecessary as long as both terms in (\ref{eq:rr_decomp}) are bounded by $O(\sqrt{T})$. 
Specially, $\textbf{Term}_1$ equals
\begin{equation}
\sum_{t=1}^T{\left(t v'^{\overline{\sigma}^t}_p - (t - 1)v'^{\overline{\sigma}^{t-1}}_p  - v^{\sigma^{t}}_p\right)}.
\end{equation}
Thanks to Lemma \ref{lm:v_eq_v_r} and Lemma \ref{lm:rec_v_eq_v_r}, for $t \geq 1$, we have 
\begin{equation}
\label{eq:tv_t_1_v_v}
\begin{aligned}
&  t v'^{\overline{\sigma}^t}_p - (t - 1)v'^{\overline{\sigma}^{t-1}}_p  - v^{\sigma^{t}}_p \\ 
= & \sum_{I\in \mathcal{I}_p}\sum_{a\in A(I)}{\pi^{\sigma^{t+1}}_p(I)g'^t_p(I, a) {\sigma}^{t+1}_p(I, a)},
\end{aligned}
\end{equation}
where $g'^t_p(I, a) = \left(R'^{t-1}_p(I, a) + r^{\sigma^t}_p(I, a)\right) - R'^{t}_p(I, a)$. We set $R'^{0}_p(I, a) = 0$ and $v'^{\overline{\sigma}^{0}}_p = 0$ as they are irrelevant to the algorithm.
Here both $R'^{t-1}_p(I, a) + r^{\sigma^t}_p(I, a)$ and $R'^{t}_p(I, a)$ are the substitute regrets of infoset $I$ at iteration $t$, according to (\ref{eq:sub_regret}). So it is potential that the difference is negligible.
Based on the above analysis, Theorem \ref{th:re_bound} is obtained. The proof is given in Appendix B, where (\ref{eq:tv_t_1_v_v}) is also proven.

\begin{theorem}
\label{th:re_bound}
After $T$ iterations of ReCFR are played, for each player $p\in P$, if $\lambda^t_p(I) > 0$ at every infoset, then, $R^T_p \leq 
\sum_{t=1}^{T}\sum_{I\in \mathcal{I}_p}\frac{1}{2\sqrt{\lambda^t_p(I)}}{\left({\lambda^{t-1}_p(I)} - {\lambda^{t}_p(I)} + \sum_{a}(r^{\sigma^t}_p(I, a))^2\right)_+} + \sum_{I\in \mathcal{I}_p}\sqrt{\lambda^T_p(I)}$.
\end{theorem}

Theorem \ref{th:re_bound} shows that the two terms in (\ref{eq:rr_decomp}) are bounded by the two terms in the theorem, respectively.
When ${\lambda^{t-1}_p(I)} - {\lambda^{t}_p(I)} + \sum_{a}(r^{\sigma^t}_p(I, a))^2 \leq 0$,
$\textbf{Term}_1$ in (\ref{eq:rr_decomp}) will be bounded by zero, and thus $\epsilon(\overline{\sigma}^T) \leq \frac{1}{T}\sum_{p\in P}\sum_{I\in \mathcal{I}_p}\sqrt{\lambda^T_p(I)}$, as shown in Corollary \ref{co:re_0}. 
\begin{corollary}
\label{co:re_0}
If $\lambda^t_p(I) = \pi^{\overline{\sigma}^t}_{-p}(I)\D^2(I)|A(I)|t$ at each infoset, then, $
\epsilon(\overline{\sigma}^T) \leq \frac{1}{\sqrt{T}}\sum_{p\in P}\sum_{I\in\mathcal{I}_p}\sqrt{\pi^{\overline{\sigma}^T}_{-p}(I)}\D(I)\sqrt{|A(I)|}$.
\end{corollary}
This corollary is based on an inequality \cite{StrategyBase}:
\begin{equation}
\label{eq:r_bound}
    {\sum_{a}(r^{\sigma^t}_p(I, a))^2} \leq \pi^{{\sigma}^t}_{-p}(I)\D^2(I)|A(I)|. 
\end{equation}
Note that $\pi^{\overline{\sigma}^t}_{-p}(I) t = \sum_{k=1}^t\pi^{{\sigma}^k}_{-p}(I)$.
As we can see, $\epsilon(\overline{\sigma}^T) = O({1}/{\sqrt{T}})$.
However, setting $\lambda^t_p(I)$ according to the corollary would perform poorly because the inequality in (\ref{eq:r_bound}) could be loose. In other words, $\lambda^t_p(I)$ could be further reduced. 
Actually, we can choose $\lambda^t_p(I)$ in a broad range.
According to Theorem \ref{th:re_bound}, the total regret is bounded by $O(\sqrt{T})$ as long as ${\lambda^{t-1}_p(I)} - {\lambda^{t}_p(I)}$ is upper bounded and $\lambda^t_p(I) = \Theta(t)$.
\begin{corollary}
\label{co:re_1}
If $\lambda^t_p(I) = \lambda\pi^{\overline{\sigma}^t}_{-p}(I)\D^2(I)|A(I)|t $, then
$\epsilon(\overline{\sigma}^T) \leq \left(\frac{1}{\sqrt{\lambda}} + \sqrt{\lambda}\right)\frac{1}{\sqrt{T}}\sum_{p\in P}\sum_{I\in\mathcal{I}_p}\sqrt{\pi^{\overline{\sigma}^T}_{-p}(I)}\D(I)\sqrt{|A(I)|}$.
\end{corollary}
The proofs for both corollaries are given in Appendix B.
In Corollary \ref{co:re_1}, $\lambda \in (0, \infty)$ is a hyper-parameter. As we can see, the optimal exploitability bound is achieved when $\lambda = 1$. 
Although Corollary \ref{co:re_1} provides a bound worse than that in Corollary \ref{co:re_0}, it allows $\lambda$ to be chosen in $(0, \infty)$.

Note that Theorem \ref{th:re_bound} does not apply when $\lambda^t_p(I) = 0$ at some infosets. Therefore, the new theoretical results do not apply to CFR and XFP.
Actually, when ReCFR is equivalent to CFR, $\textbf{Term}_1$ in  (\ref{eq:rr_decomp}) equals zero, and the known inequality $R^T_p \leq \sum_{I\in \mathcal{I}_p} \max_a (R^{T}_p(I, a))_+$ is recovered. When ReCFR is equivalent to XFP, $\textbf{Term}_2$ equals zero, and $\textbf{Term}_1$ is exactly the definition of $R^T_p$ given in (\ref{eq:total_regret}).
So, Equation (\ref{eq:rr_decomp}) does not provide new information for CFR or XFP. However, we can consider ReCFR a method that generalizes CFR and XFP.

\subsection{Adapting the hyper-parameter}
Corollary \ref{co:re_1} suggests that the optimal $\lambda$ should be $1$.
However, empirical results in Figure \ref{fig:cfr_rsv} show that the hyper-parameter could be much smaller. The reason may be that the bound for $\textbf{Term}_1$ in (\ref{eq:rr_decomp}) is too loose. So, a smaller $\lambda$ is needed to balance the two terms. 
Although Assumption \ref{ass:2} is not required in ReCFR, satisfying it may make the algorithm behave like a CFR.
In this paper, we propose to use a simple adaptive algorithm to maintain that $v'^{\overline{\sigma}^t}_1 + v'^{\overline{\sigma}^t}_2 = 0$ loosely. Note that increasing $\lambda^t_p(I)$ at any infoset could reduce $v'^{\overline{\sigma}^t}_p$. So, we check $v'^{\overline{\sigma}^t}_1 + v'^{\overline{\sigma}^t}_2$ at every iteration. If it is greater than 0, we increase $\lambda$: $\lambda = \beta_{amp} \lambda$ with $\beta_{amp} > 1$. Otherwise, $\lambda$ is reduced: $\lambda = \beta_{damp}\lambda$ with $\beta_{damp} < 1$.

\section{A new Neural Network Approximation CFR}
\label{sec:ncfrb}
This section describes Neural ReCFR with bootstrapping (Neural ReCFR-B) in detail. At each iteration, Neural ReCFR-B approximates the RSVs instead of the cumulative regrets. The RSVs are computed according to (\ref{eq:rec_sub_v}) and (\ref{eq:re_cons}).
According to (\ref{eq:rec_sub_v}), $v'^{\overline{\sigma}^t}_p(I, a)$ is scaled by the reach of the opponent and the chance player. Dividing it by ${\pi^{\overline{\sigma}^t}_{-p}(I)}$ will unify the ranges of the RSVs, which is helpful for neural approximation.
Let $u'^{\overline{\sigma}^t}_p(I, a) = {v'^{\overline{\sigma}^t}_p(I, a)}/{{\pi^{\overline{\sigma}^t}_{-p}(I)}}$
and $u'^{\overline{\sigma}^t}_p(I) = {v'^{\overline{\sigma}^t}_p(I)}/{\pi^{\overline{\sigma}^t}_{-p}(I)}$.
In this paper, we choose to approximate $u'^{\overline{\sigma}^t}_p(I, a)$ instead of $v'^{\overline{\sigma}^t}_p(I, a)$. Put $u'^{\overline{\sigma}^t}_p(I, a)$ into (\ref{eq:re_cons}), then, $u'^{\overline{\sigma}^t}_p\left(I\right)$ is required to fulfill
\begin{equation}
\label{eq:re_cons_1}
\begin{aligned}
\sum_{a\in A(I)}{\left(u'^{\overline{\sigma}^t}_p\left(I, a\right) - u'^{\overline{\sigma}^t}_p\left(I\right)\right)_+^2} 
= \beta^t_p(I),
\end{aligned}
\end{equation}
where $\beta^t_p(I) = {\lambda^t_p(I)}/{(\pi^{\overline{\sigma}^t}_{-p}(I)t)^{2}}$. According to Corollary \ref{co:re_1}, we set $\beta^t_p(I) =  {\lambda\D^2(I)|A(I)|}/{(\pi^{\overline{\sigma}^t}_{-p}(I)t)}$.
Put $u'^{\overline{\sigma}^t}_p(I, a)$ into (\ref{eq:rec_sub_v}), we get
\begin{equation}
\begin{aligned}
\label{eq:sb_q_exp}
u'^{\overline{\sigma}^t}_p(I, a) = &
    \mathbb{E}_{h \sim I, h'\sim Succ_p^{\overline{\sigma}^t}(h\cdot a)}\big\{\mathbbm{1}_{h'\in Z}u_p(h') +\\ 
    &\quad \mathbbm{1}_{h'\notin Z}u'^{\overline{\sigma}^t}_p(I(h'))\big\}.
\end{aligned}
\end{equation}
A proof for the equation is provided in Appendix C.
Equation (\ref{eq:sb_q_exp}) implies that if we \textit{sample} the payoffs and the RSVs of the earliest reachable histories starting from $(I, a)$, then, the expectation of the sampled values is precisely the RSV of action $a$ at infoset $I$. 
According to this equation, a bootstrap method similar to Q-learning \cite{QL} is derived.

\subsection{Bootstrap Learning for RSVs}
According to Equation (\ref{eq:sb_q_exp}), we propose ReCFR with bootstrapping (ReCFR-B).
\begin{define} \textbf{ReCFR-B} is an ReCFR that learns the RSVs using a bootstrap method:
at each iteration $t$ of ReCFR-B, initialize  $u'^{\overline{\sigma}^t,1}_p(I, a)$ with an arbitrary value for each  $p\in P$, $I\in \mathcal{I}_p$, and $a\in A(I)$. Let each $p\in P$ play $K$ games with the opponent who uses strategy  $\overline{\sigma}^t_{-p}$.
During the play, for each transition $(h, a, h')$ encountered in game $1 \leq k \leq K$, update $u'^{\overline{\sigma}^t,k}_p(I(h), a)$ according to
\begin{equation*}
\begin{aligned}
     & u'^{\overline{\sigma}^t,k+1}_p (I(h), a)  = u'^{\overline{\sigma}^t,k}_p(I(h), a) + \alpha_k\Big(\\
    &\quad \mathbbm{1}_{h'\in Z}u_p(h') + \mathbbm{1}_{h'\notin Z} \gamma u'^{\overline{\sigma}^t,k}_p(I(h')) - u'^{\overline{\sigma}^t,k}_p(I(h), a)\Big),
\end{aligned}
\end{equation*}
where  $u'^{\overline{\sigma}^t,k}_p\left(I\right)$ fulfills
\begin{equation*}
\sum_{a\in A(I)}{\left(u'^{\overline{\sigma}^t,k}_p\left(I, a\right) - u'^{\overline{\sigma}^t,k}_p\left(I\right)\right)_+^2} 
= \beta^t_p(I).
\end{equation*}
\end{define}

\begin{theorem}
\label{th:convq}
At each iteration $t$ of ReCFR-B, if every terminal history is visited with a non-zero probability in each game, $0 < \gamma \leq 1$, $\sum_k{\alpha_k} = \infty$, and $\sum_k{\alpha_k^2} < \infty$, then,
$u'^{\overline{\sigma}^t,K}_p(I, a)$ converges  to $u'^{\overline{\sigma}^t}_p(I, a)$ w.p.1 for every $(I, a)$ as $K \to \infty$.
\end{theorem}
The proof is provided in Appendix C. In practice, we set $\alpha_k$ to a constant learning rate and set $\gamma$ to 1 as NFSP did \cite{NFSP}.
According to the definition, the update of $u'^{\overline{\sigma}^t, k}_p(I, a)$ does not depend on how infoset $I$ is reached and how the player $p$ acts in descendants. So the learning is \textbf{off-policy}. 
ReCFR-B is similar to (batch-)OSCFR \cite{MCCFR}, as both of them update the values of infosets by sampling trajectories.
OSCFR is usually considered model-free \cite{dream,ARMAC}.\footnote{The players are regarded as a whole. However, the chance player's strategy is a part of the transition model.}
So, if $\beta^t_p(I)$ for each infoset is chosen without referring to any private information, e.g., $\sigma_{c}(h,a)$ and $\Delta(I)$, about the transition model, we can consider ReCFR-B \textbf{model-free}.
In practice, we use $\beta^t_p(I) =  {\lambda\D^2(I)|A(I)|}/{(\pi^{\overline{\sigma}^t}_{-p}(I)t)}$ as it gives a good exploitability bound.

\subsection{Neural ReCFR with Bootstrapping}

\begin{algorithm*}[t]
	\caption{Neural ReCFR with bootstrapping}
	\label{alg:NCFRB}
	\begin{algorithmic}[1]
		\State{Input: maximum iterations $T$.}
		\State{Initialize RSV network parameters $\theta^0_p$ and policy network parameters $\phi^0_p$ for each player.}
		\State{Initialize strategy memory buffer $\mathcal{M}^\Pi_p$ for each player.}
		\For {iteration $t= 1$ to $T$}
        	\For {player $p\in P$}
        	    \State Initialize a temporary data set $\mathcal{D}^\mathcal{R}_p$.
        	    \For {game $k = 1$ to $K$}
        	        \State{Sample a strategy $\hat{\sigma}_i$ for each player $i\in P$,
        	        $\hat{\sigma}_i = 
        	            \begin{cases}
        	                \sigma^t_i, &\text{ with probability } \eta,\\
        	                \Pi\left(\phi^{t - 1 }_i\right), &\text{ with probability } 1 - \eta.
        	            \end{cases}$
        	        }
                	\State $\mathcal{T}^k_p \gets \operatorname{Play}(\hat{\sigma}_p, \hat{\sigma}_{-p})$. \Comment{Sample a trajectory $\mathcal{T}^k_p$ using $\hat{\sigma}$}.
                    \State{Collect all transitions $\left(h, a, h', \pi^{\overline{\sigma}^t}_{-p}(I(h))\right) \in \mathcal{T}^k_p$  to data set $\mathcal{D}^{\mathcal{R}}_p$}.
                     \If {player $p$ follows the current strategy $\sigma^t$}
                            \State{Store all behavior tuples $(h, a) \in \mathcal{T}^k_p$ in strategy memory buffer $\mathcal{M}^\Pi_p$ }.
                    \EndIf
                 \EndFor
                  \State Train $\theta^t_p$ on loss (\ref{eq:loss_r}) and train $\phi^t_p$ on loss (\ref{eq:loss_pi}).
            \EndFor
            
         \EndFor
	\end{algorithmic}
\end{algorithm*}

For each player, an RSV network $\mathcal{R}(\theta^t_p)$ with parameters $\theta^t_p$ is used to approximate $u'^{\overline{\sigma}^t}_p(I, a)$ at all infosets. Also, a neural network $\Pi(\phi^t_p)$ with parameters $\phi^t_p$ is used to approximate the average strategy $\overline{\sigma}^t_p$. 
At iteration $t$, the next iteration strategy $\sigma^{t+1}_p$ is computed according to the output of the RSV network,
\begin{equation}
\sigma^{t+1}_p(I, a) = \operatorname{RM}\left(\mathcal{R}\left(I, a|\theta^{t}_{p}\right) - u'^{\overline{\sigma}^{t}}_{p}(I)\right),
\end{equation}
where $u'^{\overline{\sigma}^{t}}_{p}(I)$ is the RSV of infoset $I$, and it is chosen to fulfill (\ref{eq:re_cons_1}) with $u'^{\overline{\sigma}^{t}}_{p}(I, a) \gets \mathcal{R}\left(I, a|\theta^{t}_{p}\right)$. Since RM is \textit{scale-invariant}, this equation is equivalent to (\ref{eq:re_st}) if the approximation is accurate.
We use anticipatory dynamics \cite{NFSP} to estimate the average strategy $\overline{\sigma}^t$. Specifically, the strategy for each player at iteration $t$ is 
$
\hat{\sigma}_p = (1 - \eta)\Pi\left(\phi^{t - 1 }_p\right) + \eta \sigma^{t}_p
$,
where $\eta$ is a hyper-parameter.
We set $\eta = 0.1$ as NFSP did. In practice, multiple trajectories are sampled. So, $\Pi\left(\phi^{t - 1 }_p\right)$ is selected with probability $1 - \eta$ for playing, while $\sigma^{t}_p$ is selected with probability $\eta$.
The pseudocode of Neural ReCFR-B is given in Algorithm \ref{alg:NCFRB}. 
At each iteration and for each player $p$, $K$ games are played using the anticipatory strategy $\hat{\sigma}$. During the play, the transitions are collected into a temporary data set $\mathcal{D}^{\mathcal{R}}_p$, and the behavior tuples are stored in a reservoir buffer \cite{NFSP} $\mathcal{M}^\Pi_p$. The RSV network for player $p$  is trained by minimizing:
\begin{equation}
\label{eq:loss_r}
\begin{aligned}
\mathcal{L}(\theta^t_p)  =  & \mathbb{E}_{\left(h, a, h', \pi^{\overline{\sigma}^t}_{-p}(I(h))\right) \sim \mathcal{D}^{\mathcal{R}}_p}\big\{\big(\mathbbm{1}_{h'\in Z}u_p(h') + \\ 
& \quad \mathbbm{1}_{h'\notin Z}u'^{\overline{\sigma}^{t}}_p(I(h')) - \mathcal{R}(I(h), a|\theta^t_p)\big)^2\big\},
\end{aligned}
\end{equation}
where $\theta^t_p$ is the trainable parameters initialized with $\theta^{t-1}_p$, and $u'^{\overline{\sigma}^{t}}_p(I(h'))$ is chosen to fulfill (\ref{eq:re_cons_1}) with $u'^{\overline{\sigma}^{t}}_p(I(h'), a) \gets \mathcal{R}\left(I(h'), a|\theta^t_p\right)$.
The average network for player $p$ is trained by minimizing the cross-entropy loss:
\begin{equation}
\label{eq:loss_pi}
\mathcal{L}(\phi^t_p) = \mathbb{E}_{(h, a) \sim \mathcal{M}^\Pi_p}\left\{-\log \Pi(I(h), a|\phi^t_p)\right\},
\end{equation}
where $\phi^t_p$ is the trainable parameters initialized with $\phi^{t-1}_p$.

The algorithm requires each player $p\in P$ plays with an opponent who uses the average strategy $\overline{\sigma}^{t}_{-p}$. However, the strategy for player $p$ is not specified. We propose \textbf{asymmetric learning},  in which player $p$ uses a mixed strategy of the uniform random strategy and strategy $\sigma^{t}_p$, i.e.,  $\hat{\sigma}_p = (1 - \eta)\operatorname{Uniform} + \eta \sigma^{t}_p$. Accordingly, the method that both players use the anticipatory strategy is called \textbf{symmetric learning}. We use asymmetric learning in our experiments by default.

\subsection{Discussion}
\textbf{Approximating the average strategy}.
In Neural ReCFR-B, approximating cumulative regrets is avoided, but the average strategy is still approximated. Single Deep CFR \cite{SingleDeepCFR} proposed remembering all past regret networks to avoid doing that. This method should also be compatible with Neural ReCFR-B. Nevertheless, it should be easier to train the average networks since the average strategy changes more slowly than the cumulative regrets.

\textbf{Relationship with DNCFR}.
Both Neural ReCFR-B and DNCFR use bootstrap learning to reduce the variance in training targets. However, a critical difference is that Neural ReCFR-B bootstraps on the RSVs of the earliest reachable infosets, while DNCFR bootstraps on the last approximations.
Since we only consider depth-limited games, the approximation error in Neural ReCFR-B is \textit{bounded} at every iteration, while the error in DNCFR is \textit{accumulated}.

\textbf{Relationship with NFSP}.
Since ReCFR is equivalent to XFP according to Proposition  \ref{prop:re_eq_XFP} when $\lambda^t_p(I) = 0$ at all infosets, Neural ReCFR-B can also be regarded as NFSP in this case. 
Note that the learning algorithm, i.e., DQN \cite{DQN}, in NFSP has two essential components: 1) a value memory buffer, which may improve sample efficiency. 2) a target network, which may stabilize the training. We will test Neural ReCFR-B with these two components in section \ref{sec:setup}.

\section{Experimental setup and results}
\label{sec:setup}
We first test ReCFR on Leduc Poker \cite{Leduc} to show the convergence properties of the algorithm. Leduc poker is a small size game with two rounds of betting.  Then, Neural ReCFR-B is tested on heads-up flop hold’em poker (FHP) \cite{DeepCFR} and heads-up limit Texas hold’em (HULH).\footnote{Source code: https://github.com/Liuweiming/Neural\_ReCFR\_B} FHP is a medium-size game with over $10^{12}$ nodes and $10^9$  {\infoset}s, while HULH is a large-size game with over $10^{17}$ nodes and $10^{14}$ infosets. More details about the games are given in Appendix D.
The neural network architecture is the same as in \cite{DeepCFR}. It is a seven-layer fully connected neural network.

We train the neural networks using Adam optimizer \cite{adam}, with a batch size of 128, a learning rate of 0.001, and gradient norm clipping to 1. At every iteration, 1000 plays are performed to collect samples. The RSV networks are trained for two epochs (approximately 32 SGD steps with a batch size of 128 on FHP), while the average networks are trained for 16 SGD steps.
The strategy memory size is set to 10 million.  For HULH, we increase the batch size to 6,400, the number of SGD steps to 64, and the number of plays to 100,000. The strategy memory size is also increased to 40 million. We also test Neural ReCFR-B with two RSV memory buffers (one for each player). In this setting, the RSV memory sizes for FHP and HULH are 1 million and 4 million, respectively, and the RSV networks are trained for 32 and 64 SGD steps, respectively.

Neural ReCFR-B is compared with two model-based neural CFRs: Deep CFR and DNCFR; and two model-free algorithms: DREAM
and NFSP, on FHP. A model-free variant of Deep CFR with Outcome Sampling \cite{MCCFR} (Deep OSCFR) is also included. All the algorithms are implemented based on OpenSpiel \cite{openspiel}.
We implement Deep CFR and DREAM with the hyper-parameters given in \cite{DeepCFR} and \cite{dream}, respectively.
The hyper-parameters for DNCFR and NFSP are determined through a set of experiments. A head-to-head comparison is performed between Neural ReCFR-B and Deep CFR on HULH. All the settings are given in Appendix E.

Performance is measured using exploitability in terms of milli big blinds per game (mbb/g).
For a specific exploitability value, the numbers of nodes touched and the numbers of samples consumed ($=$ \#SGD steps $\times$ batch size) of different algorithms are compared, respectively. Algorithms touching fewer nodes are more sample efficient, while algorithms consuming fewer samples are more training efficient.

\subsection{Experimental Results of ReCFR}
\label{sec:emp_result}
\begin{figure}[t]
\centering
\begin{subfigure}[b]{0.493\linewidth}
    \centering
    \includegraphics[width=\linewidth]{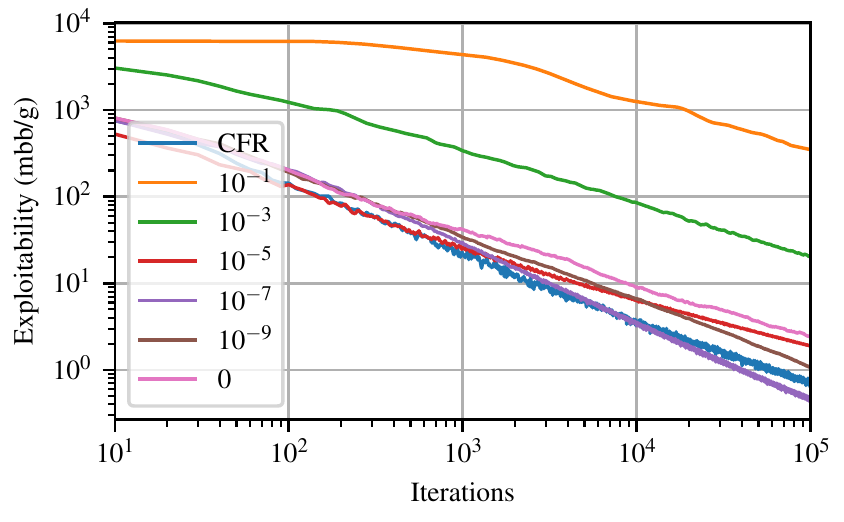}
    \vspace*{-0.25in}
    \label{fig:cfr_rsv_no_adapt_0}
\end{subfigure}
\begin{subfigure}[b]{0.493\linewidth}
    \centering
    \includegraphics[width=\linewidth]{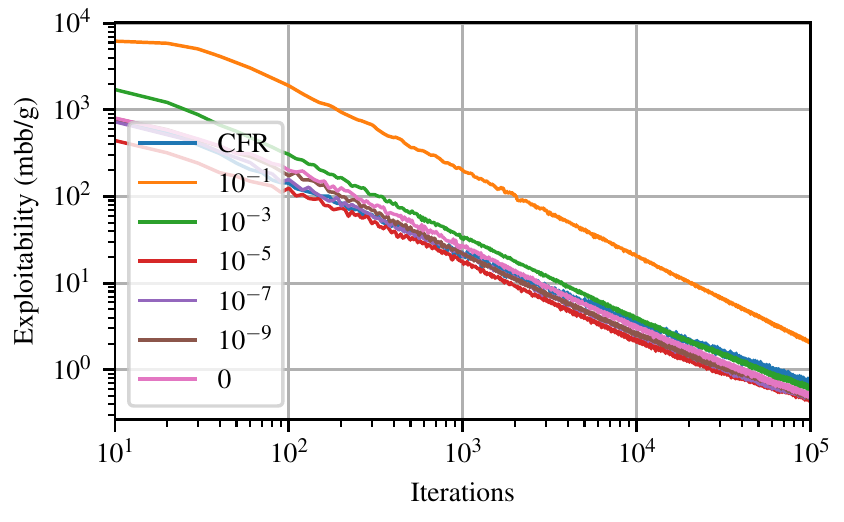}
    \vspace*{-0.25in}
    \label{fig:cfr_rsv_adapt_0}
\end{subfigure}
    \caption{Exploitability curves of vanilla CFR and ReCFR on Leduc poker. \textbf{Left}: ReCFR with constant $\lambda$s shown in the legend. \textbf{Right}: ReCFR with adaptive $\lambda$s initialized with the values in the legend.}
    \label{fig:cfr_rsv}
\end{figure}

In this subsection, we test ReCFR on Leduc poker \cite{Leduc}. 
ReCFR is compared with vanilla CFR, as the latter can be regarded as a special case of ReCFR.
We first test ReCFR with a constant $\lambda$ in $[0, 1]$. As shown on the left side in Figure \ref{fig:cfr_rsv}, ReCFR with any $\lambda$ converges. It seems ReCFR with $\lambda = 10^{-7}$ is the fastest, even faster than vanilla CFR.
However, if $\lambda$ is reduced to $0$, ReCFR will degenerate to XFP, and it converges slower than CFR. We also test ReCFR with an adaptive $\lambda$, as shown on the right side in Figure \ref{fig:cfr_rsv}. As we can see, ReCFR with different initial $\lambda$ (even 0) converges as fast as CFR, except the one with an initial $\lambda = 1$. 
So, with the adaptive method, it is easier to choose the hyper-parameter for ReCFR, as long as the initial value is small enough. In Appendix F, the curves of the adaptive $\lambda$ in ReCFR with different initial values are given. 

\subsection{Experimental Results of Neural ReCFR-B}
\label{sec:result_cfr}

\begin{figure}[t]
    \centering
    \begin{subfigure}[b]{0.493\linewidth}
        \centering
        \includegraphics[width=\linewidth]{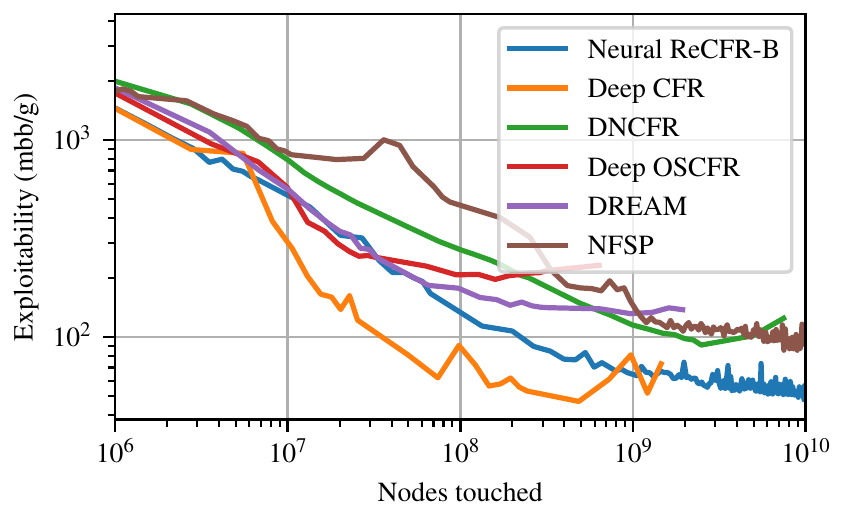}
        \vspace*{-0.25in}
        \label{fig:plot_node_0}
    \end{subfigure}
    \hfill
    \begin{subfigure}[b]{0.493\linewidth}
        \centering
        \includegraphics[width=\linewidth]{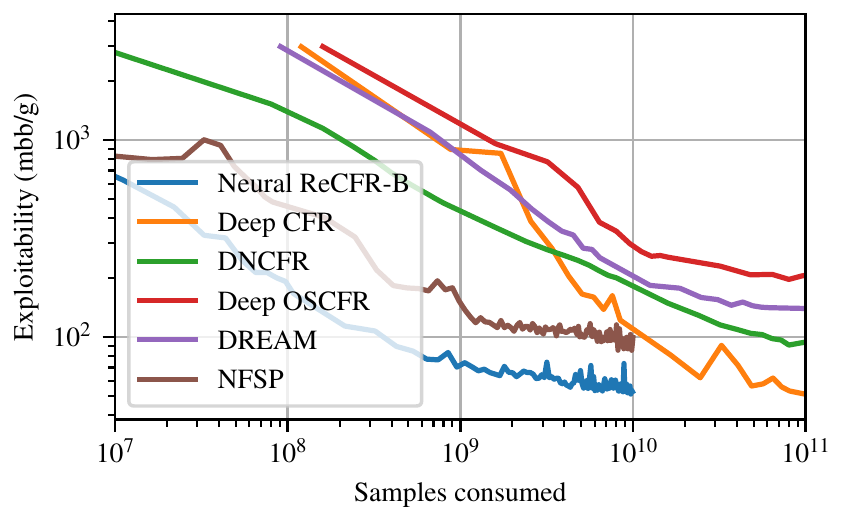}
        \vspace*{-0.25in}
        \label{fig:plot_sample_0}
    \end{subfigure}
    \caption{Exploitability curves of different algorithms on FHP. The x-axes represent the number of nodes touched and the number of samples consumed ($=$ \#SGD steps $\times$ batch size), respectively.}
    \label{fig:comp_all}
\end{figure}

The results of all the algorithms on FHP are given in Figure \ref{fig:comp_all}. As we can see, Neural ReCFR-B achieves an exploitability of 50.5 mbb/g after touching $1.0 \times 10^{10}$ nodes, while Deep CFR achieves 47.0 mbb/g after touching $6.0 \times 10^8$ nodes, but the value increases to 73.0 mbb/g after touching $1.3 \times 10^9$ nodes. Neural ReCFR-B is worse than Deep CFR in sample efficiency. It is reasonable because model-free algorithms are less sample efficient by nature. However, our algorithm is faster than the three model-free algorithms: Deep OSCFR, DREAM, and NFSP. More importantly, our algorithm is the most training efficient, by more than \textbf{25} times faster than Deep CFR to reach the exploitability of 100 mbb/g, while Deep OSCFR and DREAM never reach this value and both DNCFR and NFSP are stuck at this value. Recall that DNCFR also uses bootstrap learning. However, it has a higher exploitability than Neural ReCFR-B. On the other hand, NFSP is training efficient, but it achieves an exploitability much higher than Neural ReCFR-B and Deep CFR. Note that NFSP can be regarded as a special case of Neural ReCFR-B to some extent, and the latter may degenerate to NFSP when the gap between $u'^{\overline{\sigma}^t}_p(I)$ and $\max_{a} u'^{\overline{\sigma}^t}_p(I, a)$ at every infoset is too small to be learned by neural networks. As for DREAM, it performs better than Deep OSCFR but worse than Neural ReCFR-B and Deep CFR. This suggests that DREAM is still suffering from high variance training targets.

\begin{figure}[t]
\centering
    \begin{subfigure}[b]{0.493\linewidth}
        \centering
        \includegraphics[width=\linewidth]{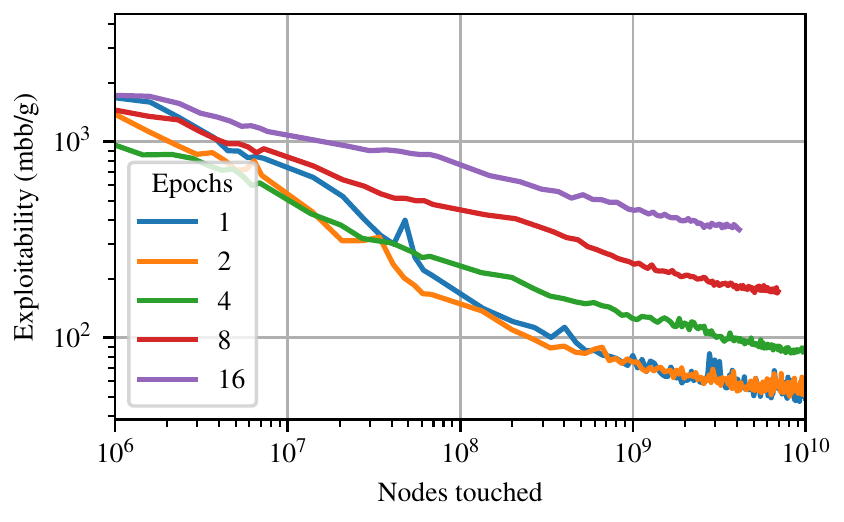}
        \vspace*{-0.25in}
    \end{subfigure}
    \hfill
    \begin{subfigure}[b]{0.493\linewidth}
        \centering
        \includegraphics[width=\linewidth]{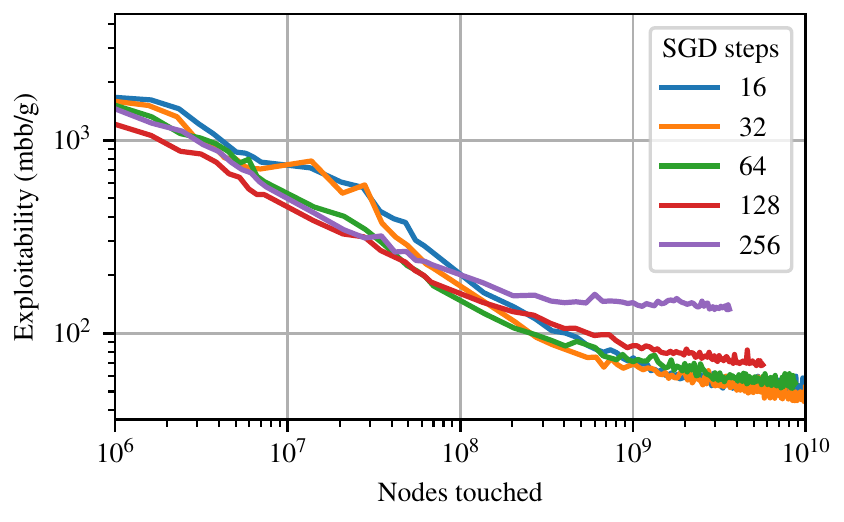}
        \vspace*{-0.25in}
    \end{subfigure}
    \caption{Exploitability curves of Neural ReCFR-B with different numbers of epochs / SGD steps on FHP. \textbf{Left}: the default Neural ReCFR-B.  \textbf{Right}: Neural ReCFR-B with RSV memory buffers.}
    \label{fig:epoch}
\end{figure}

\begin{figure}[t]
\centering
    \begin{subfigure}[b]{0.493\linewidth}
        \centering
        \includegraphics[width=\linewidth]{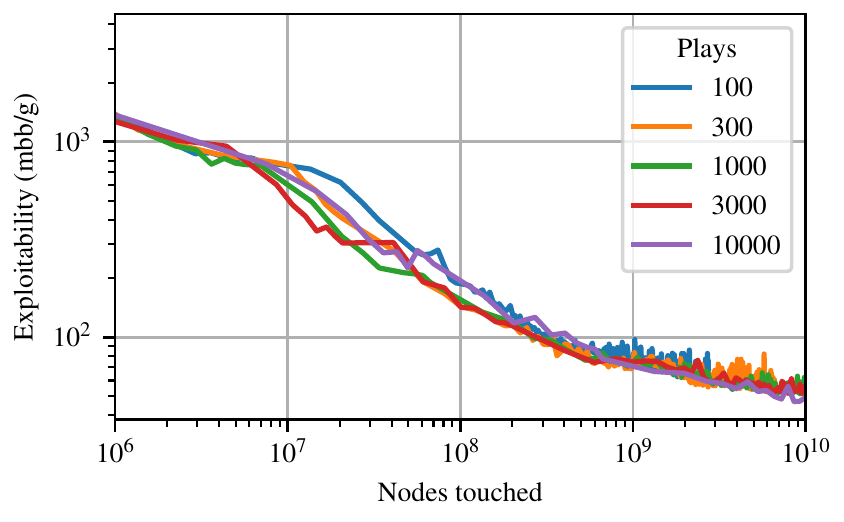}
        \vspace*{-0.25in}
    \end{subfigure}
    \hfill
    \begin{subfigure}[b]{0.493\linewidth}
        \centering
        \includegraphics[width=\linewidth]{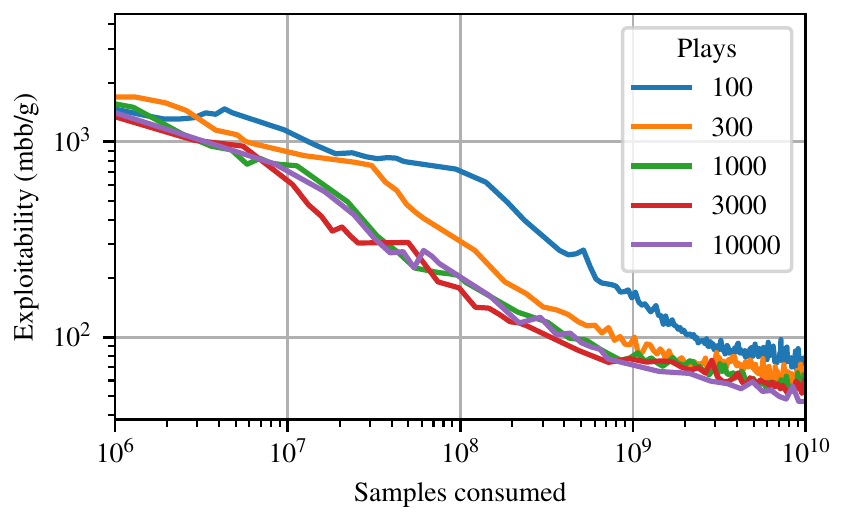}
        \vspace*{-0.25in}
    \end{subfigure}
    \caption{Exploitability curves of Neural ReCFR-B with different numbers of plays on FHP. The x-axes represent the number of nodes touched and the number of samples consumed, respectively.}
    \label{fig:cfr}
\end{figure}

We also tried to improve the training efficiency of Deep CFR by only reducing the number of SGD steps for training the regret networks. The results in Appendix F show that Deep CFRs with fewer SGD steps converge earlier to higher exploitability. 
On the other hand, as shown in Figure \ref{fig:epoch}, Neural ReCFR-B converges to a similar exploitability even the RSV networks are trained for one epoch per iteration. 
To further investigate the robustness of Neural ReCFR-B, we test it with different settings on FHP. In Figure \ref{fig:epoch}, the results of Neural ReCFR-B with different numbers of epochs for training the RSV networks are given.
As we can see, training the RSV networks for more than two epochs can reduce the convergence speed significantly. A possible reason is that the neural networks are overfitting. We conjecture that using memory buffers can partially solve this problem. As shown on the right side in Figure \ref{fig:epoch}, the problem is alleviated. Overfitting is also observed in Deep CFR \cite{DeepCFR}, and it is solved by training the regret networks from scratch at every iteration. In Neural ReCFR-B, reducing the number of epochs to alleviate the problem is more appealing. 
In Figure \ref{fig:cfr}, the results of Neural ReCFR-B with different numbers of plays per iteration are given. It is shown that Neural ReCFR-B is insensitive to the number of plays in sample efficiency. However, performing more plays (collecting more training data) per iteration can increase the training efficiency. This is because of the variance in approximated RSVs is also affected by the data set size.

\begin{figure}[t]
    \centering
    \begin{subfigure}[b]{0.493\linewidth}
        \centering
        \includegraphics[width=\linewidth]{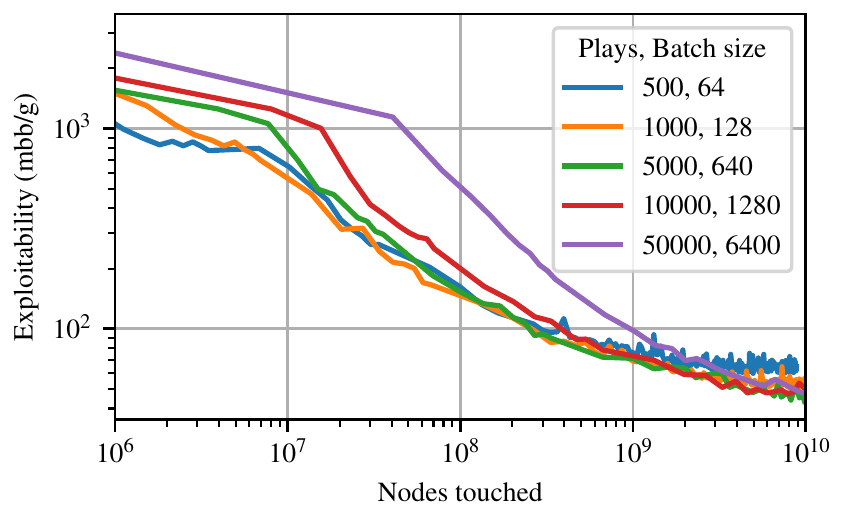}
        \vspace*{-0.25in}
    \end{subfigure}
    \hfill
    \begin{subfigure}[b]{0.493\linewidth}
        \centering
        \includegraphics[width=\linewidth]{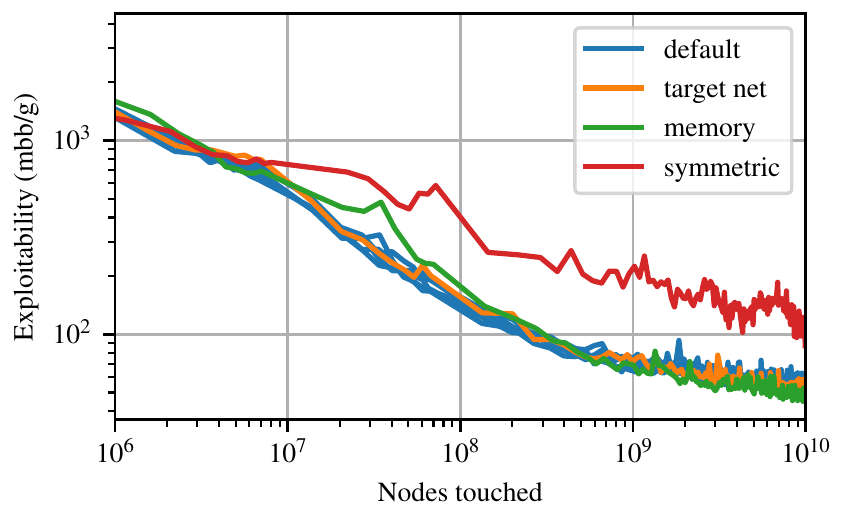}
        \vspace*{-0.25in}
    \end{subfigure}
    \caption{ \textbf{Left}: Exploitability curves of Neural ReCFR-B with different numbers of plays and batch sizes. \textbf{Right}: Exploitability curves of Neural ReCFR-B with additional components. We also plot the exploitability curves of five replicates of Neural ReCFR-B (the blue lines) with different random seeds.}
    \label{fig:comp_pow_1_cfr}
\end{figure}

On the left side in Figure \ref{fig:comp_pow_1_cfr}, we test the scalability of the algorithm by increasing the number of plays per iteration and the batch size simultaneously.
As we can see, the algorithm can scale down to 500 plays and up to 50,000 plays per iteration. Performing more plays per iteration and using a larger batch size may reduce the sample efficiency. However, it is easier to parallelize and thus may reduce the training time dramatically. In conclusion, Neural ReCFR-B is robust to different hyper-parameters.

On the right side in Figure \ref{fig:comp_pow_1_cfr}, we test Neural ReCFR-B with additional components. The default is Neural ReCFR-B with asymmetric learning, without memory buffers and target networks. 
As we can see, Neural ReCFR-B with target networks performs identically to the default one. On the other hand, Neural ReCFR-B with RSV memory buffers seems to have a slightly lower exploitability. This result suggests that using RSV memory buffers in Neural ReCFR-B may increase the sample efficiency. 
However, Neural ReCFR-B with symmetric learning is worse than the default one. The reason might be that symmetric learning can not guarantee to visit every infoset with a probability greater than zero.
Besides, we plot five replicates of Neural ReCFR-B with different random seeds (so the neural networks are also initialized differently). As we can see, the results are consistent.

\begin{figure}[t]
	\centering
    \begin{subfigure}[b]{0.493\linewidth}
        \centering
        \includegraphics[width=\linewidth]{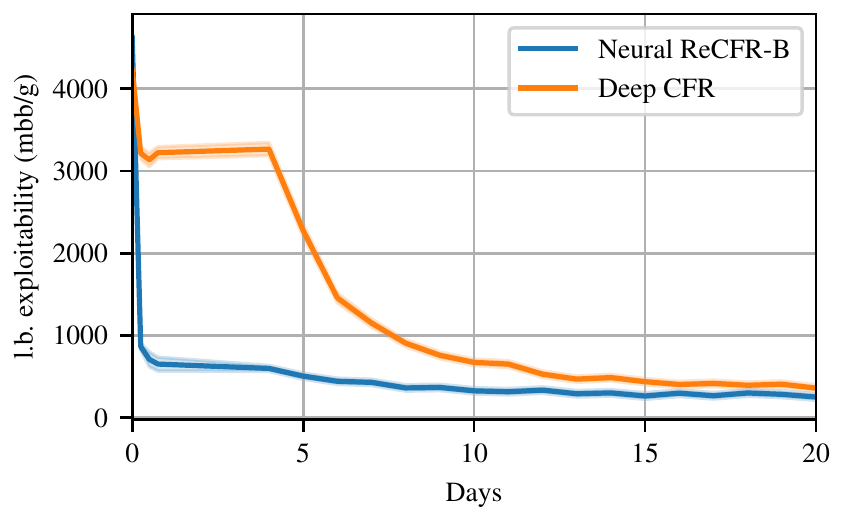}
        \vspace*{-0.25in}
    \end{subfigure}
    \hfill
    \begin{subfigure}[b]{0.493\linewidth}
        \centering
        \includegraphics[width=\linewidth]{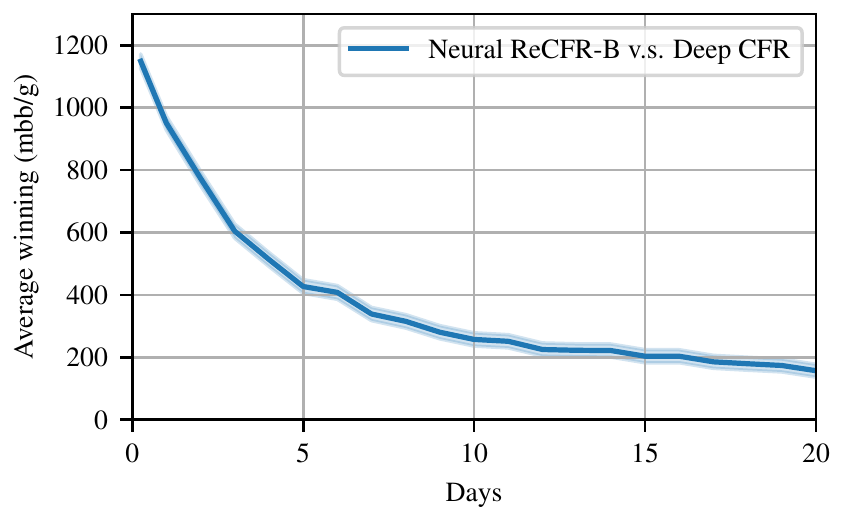}
        \vspace*{-0.25in}
    \end{subfigure}
    \caption{Comparison of Neural ReCFR-B and Deep CFR on HULH. The x-axis indicates the days for training.  \textbf{Left}: Lower bound on exploitability. \textbf{Right}: Head-to-head performance. The y-axis represents the average winning of Neural ReCFR-B (simulated by $10^6$ plays). All the values are presented with 95\% confidence interval.} 
    \label{fig:match}
\end{figure}

\begin{figure}[t]
	\centering
    \begin{subfigure}[b]{0.493\linewidth}
        \centering
        \includegraphics[width=\linewidth]{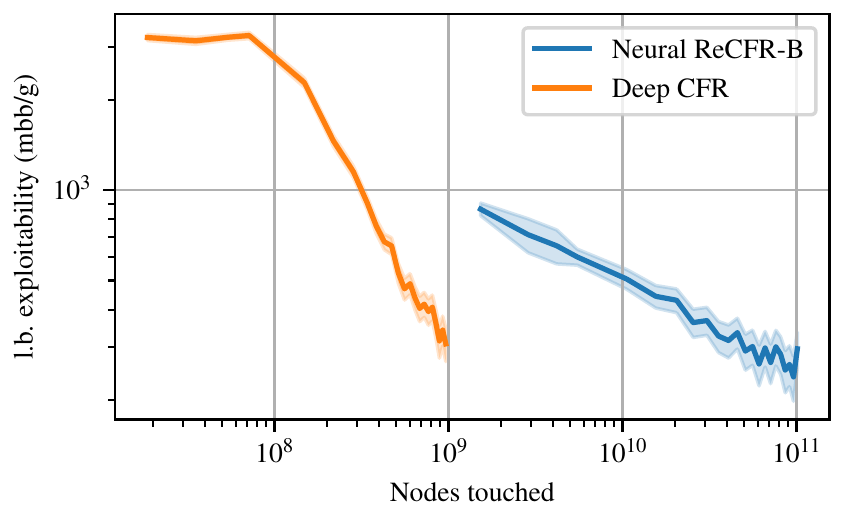}
        \vspace*{-0.25in}
    \end{subfigure}
    \hfill
    \begin{subfigure}[b]{0.493\linewidth}
        \centering
        \includegraphics[width=\linewidth]{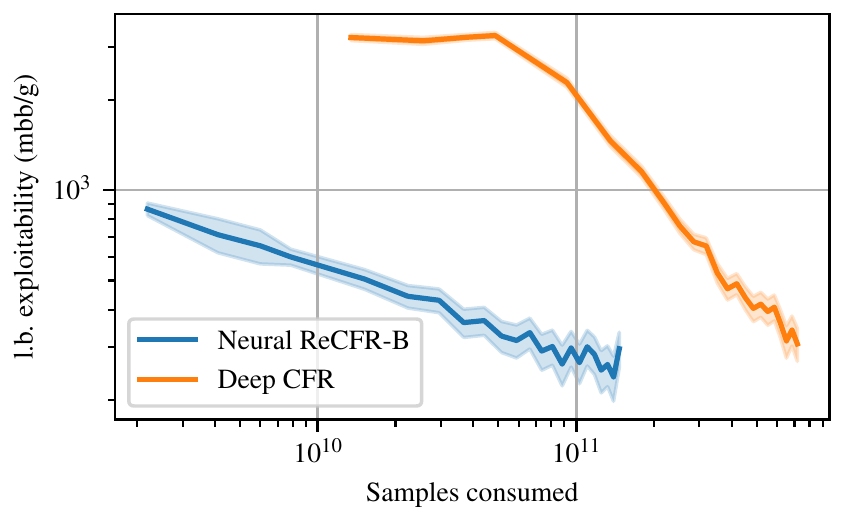}
        \vspace*{-0.25in}
    \end{subfigure}
    \caption{Curves of lower bounds on exploitability of Neural ReCFR-B and Deep CFR on HULH. The x-axes represent the number of nodes touched and the number of samples consumed, respectively.}
    \label{fig:HuLH}
\end{figure}

Finally, we evaluate Neural ReCFR-B on HULH.
Since using RSV memory buffers may increase sample efficiency and avoid overfitting, we use RSV memory buffers in this experiment. We train Neural ReCFR-B and Deep CFR separately for 20 days on one GPU and 10 CPU cores. 
On the left side of Figure \ref{fig:match}, we compare the lower bounds on the exploitability of Neural ReCFR-B and Deep CFR. The lower bounds are estimated using a Local Best Response \cite{LBR} algorithm.
The comparison is performed day by day. So it should reflect both sample efficiency and training efficiency. 
As we can see, Neural ReCFR-B achieves a lower bound around 600 mbb/g only after one day's training, while Deep CFR reaches this value on the ninth day. After 20 days' training, Deep CFR and Neural ReCFR-B have similar lower bounds on exploitability, about 300 mbb/g. 
Besides, Figure \ref{fig:HuLH} shows that Neural ReCFR-B is more training efficient, but Deep CFR has a better sample efficiency.
Note that the sample efficiency of Neural ReCFR-B may be increased when the batch size is reduced at the cost of increasing training time, according to the scalability experiment on FHP.
On the right side of Figure \ref{fig:match}, we show the head-to-head performance between Neural ReCFR-B and Deep CFR. As we can see, Neural ReCFR-B beats Deep CFR by more than 150 mbb/g.
It looks like that both algorithms have not converged yet, but it is safe to conclude that Neural ReCFR-B is more training efficient and time-efficient.

\section{Conclusion}
\label{sec:conclusion}
This paper proposes a new CFR algorithm, ReCFR, and a model-free Neural CFR algorithm, Neural ReCFR-B. In these two algorithms, cumulative regrets are replaced by RSVs proposed in \cite{StrategyBase}. After revisiting RSVs and Warm CFR, we prove that ReCFR can converge to a Nash equilibrium at a rate of $O({1}/{\sqrt{T}})$. Thanks to the recursive and non-cumulative properties of the RSVs, when bootstrap learning is used, the variance in training targets in Neural ReCFR-B should be low. The experimental results show that Neural ReCFR-B achieves competitive performance to the state-of-the-art neural CFR algorithms with higher training efficiency.  

According to the theoretical results, it is promising to transform other CFR algorithms, e.g., CFR+ \cite{CFRPlus} and PCFR \cite{Blackwell}, to new algorithms similar to ReCFR, and develop new neural CFR algorithms based on them. Also, combining Neural ReCFR-B with variance-reduction techniques \cite{VRMCCFR,lowCFR} or other improvements \cite{SingleDeepCFR,dream,ARMAC} may produce more efficient algorithms. Furthermore, 
there are some other equilibrium-finding algorithms that have fast convergence, e.g., Excessive Gap Technique (EGT) \cite{EGT} and Optimistic Follow the Regularized Lead (OFTRL) \cite{OptimisticRegret}. And EGT does not even use any cumulative variables. However, it is unclear whether they are compatible with sampling and function approximation. So, more research is required in this direction.
Besides, it has been shown in \cite{uniview} that RM is equivalent to Follow the Regularized Lead (FTRL) \cite{FTRL}. In this paper, it is shown that ReCFR is a generalization of CFR and XFP. So, there might be some strong connections between CFR, XFP, and FTRL worthy of further study.

With the model-free Neural ReCFR-B, we may also apply CFR algorithms to a broader range of IIGs, for example, non-stationary games,  non-zero-sum games, multi-player games, and even video games.

\clearpage

\bibliographystyle{IEEEtran}
\bibliography{IEEEabrv,TOG}

\renewcommand{\thetheorem}{A.\arabic{theorem}}
\renewcommand{\thelemma}{A.\arabic{lemma}}
\renewcommand{\thedefine}{A.\arabic{define}}
\renewcommand{\thecorollary}{A.\arabic{corollary}}
\renewcommand{\theequation}{A.\arabic{equation}}

\section{Proofs and More Details for CFR and RSVs}
\subsection{Proof for Equation \ref{eq:rec_v}}

\begin{proof}
First, for any strategy $\sigma$, a history of player $p$, a descendant $h'\in Succ_p(h \cdotp a)$, and any terminal history $z\in Z$ reachable from $h$ and $h'$, according to the definition of \textit{reach}, we have
\begin{equation}
\begin{aligned}
    \pi^{\sigma}(h \cdotp a, z) = & \pi^{\sigma}_{-p}(h \cdotp a, z)\times \pi^{\sigma}_p(h \cdotp a, z) \\
    = &\pi^{\sigma}_{-p}(h \cdotp a, h')\pi^{\sigma}_{-p}(h', z)\times \pi^{\sigma}_p(h', z) \\
    = &\pi^{\sigma}_{-p}(h \cdotp a, h')\pi^{\sigma}(h', z).
\end{aligned}
\end{equation}
Note that $\pi^{\sigma}_p(h \cdotp a, z) = \pi^{\sigma}_p(h', z)$ according to the definition of $Succ_p$.
Therefore, 
\begin{equation}
\begin{aligned}
\pi^{\sigma}_{-p}(h)\pi^{\sigma}(h \cdotp a, z) = & \pi^{\sigma}_{-p}(h)\pi^{\sigma}_{-p}(h \cdotp a, h')\pi^{\sigma}(h', z) \\
= & \pi^{\sigma}_{-p}(h')\pi^{\sigma}(h', z).
\end{aligned}
\end{equation}
So, for any strategy $\sigma$, $I\in \mathcal{I}_p$ and $a\in A(I)$,
\begin{equation}
\begin{aligned}
& v^{\sigma}_p(I, a) \\
= & \sum_{h\in I}\sum_{z\in Z:h\sqsubset z}\pi^{\sigma}_{-p}(h)\pi^{\sigma}(h \cdotp a, z)u_p(z) \\
=  & \sum_{h\in I}{\sum_{z\in Z: z \in Succ_p(h \cdotp a)}{\pi_{-p}^{\sigma}(z)\pi^{\sigma}(z, z)u_p(z)}} + \\
& \sum_{h\in I}{\sum_{h'\notin Z:h' \in Succ_p(h \cdotp a)}{\sum_{z\in Z:h'\sqsubset z}\pi_{-p}^{\sigma}(h')\pi^{\sigma}(h', z)u_p(z)}} \\
= & \sum_{h\in I}{\sum_{z\in Z: z \in Succ_p(h \cdotp a)}{\pi_{-p}^{\sigma}(z)u_p(z)}} + \\
& \sum_{I'\in Succ_p(I, a)}{\sum_{h'\in I'}{\sum_{z\in Z:h'\sqsubset z}\pi_{-p}^{\sigma}(h')\pi^{\sigma}(h', z)u_p(z)}} \\
= & \sum_{h\in I}{\sum_{z\in Z: z \in Succ_p(h \cdotp a)}{\pi_{-p}^{\sigma}(z)u_p(z)}} + \\
& \sum_{I'\in Succ_p(I, a)}{v^{\sigma}_p(I')}.
\end{aligned}
\end{equation}
Note that we only consider perfect-recall games.
The third equality holds because $\{h' | h' \notin Z,h'\in Succ_p(h \cdotp a), h\in I\} = \{h' | h' \in I', I' \in Succ_p(I, a)\}$. Then, the fourth equality holds according to the definition of counterfactual values.
\end{proof}

\subsection{A Method for Computing the RSVs}
\label{sec:cfrmax_exist}
When computing the RSVs, we need to solve $v'^{\sigma}_p(I) \in \mathbb{R}$ in constraint
\begin{equation}
\label{e:eq:sb_cons}
\begin{aligned}
\quad \sum_{a\in A(I)}{\left(v'^{\sigma}_p\left(I, a\right) - v'^{\sigma}_p\left(I\right)\right)_+^2} \leq  \lambda(I),
\end{aligned}
\end{equation}
where $\lambda(I) \in (0, \infty)$ is a parameter.
There could be multiple solutions that satisfy the constraint. Define $f: v'^{\sigma}_p\left(I\right) \mapsto \sum_{a\in A(I)}{\left( v'^{\sigma}_p\left(I, a\right) - v'^{\sigma}_p\left(I\right)\right)_+^2}$.  
The constraint can be rewritten as $f(v'^{\sigma}_p(I)) \leq \lambda(I)$. To solve the inequation, we need to first solve $v'^{\sigma}_p(I)$ in equation 
\begin{equation}
\label{eq:sb_cons_eq}
    f(v'^{\sigma}_p(I)) = \lambda(I).
\end{equation}
We plotted $f(x)$ in Figure \ref{fig:app_exi_cfrmax}, in the case that $v'^{\sigma}_p\left(I, \cdot\right) = [-0.7, 0, 1]$.
As we can see, function $f(x)$ in range $(-\infty, \max_a v'^{\sigma}_p\left(I, a\right)]$ is strongly convex and strictly decreasing. 
In other words, function $f: (-\infty, \max_a v'^{\sigma}_p\left(I, a\right)]\to [0, \infty)$ is a bijection. Therefore, the solution for (\ref{eq:sb_cons_eq}) exists and is unique when $\lambda(I) > 0$.

\begin{figure}[htbp]
    \centering
    \resizebox{\linewidth}{!}{\input{images/cfr_max.pgf}}
    \vspace*{-0.25in}
    \caption{Curves of $f(x)$ when $ v'^{\sigma}_p\left(I, \cdot\right) = [-0.7, 0, 1]$.}
    \label{fig:app_exi_cfrmax}
\end{figure}

To solve (\ref{eq:sb_cons_eq}), we can first sort $ v'^{\sigma}_p\left(I, \cdot\right)$ such that $ v'^{\sigma}_p\left(I, a_1\right)\leq \dots \leq  v'^{\sigma}_p\left(I, a_{|A(I)|}\right)$. Then we can try to solve $v'^{\sigma}_p(I)$ in each interval: $\left(\infty,v'^{\sigma}_p\left(I, a_1\right)\right]$, $\dots$, $\left(v'^{\sigma}_p\left(I, a_{|A(I)|-1}\right), v'^{\sigma}_p\left(I, a_{|A(I)|}\right)\right]$. In each interval, the hard thresholding operator $(\cdot)_+$ can be removed, i.e., (\ref{eq:sb_cons_eq}) is equivalent to a conventional quadratic equation.

Since the complexity for sorting $ v'^{\sigma}_p\left(I, \cdot\right)$ and iteratively solving the quadratic equations is $O(|A(I)|^2)$, 
The complexity for solving (\ref{e:eq:sb_cons}) or (\ref{eq:sb_cons_eq}) is $O(|A(I)|^2)$. 
A similar problem can be found in \cite{Condat16}, where we may find an algorithm to solve (\ref{e:eq:sb_cons}) or (\ref{eq:sb_cons_eq}) more efficiently.


\subsection{Proof for Lemma \ref{lm:v_eq_v_r}}

\begin{proof}
The right hand side of the equation is
\begin{equation}
\begin{aligned}
& \sum_{I\in \mathcal{I}_p}\sum_{a\in A(I)}{\pi^{\sigma'}_p(I)\left(v^{\sigma}_p(I, a) - v^{\sigma}_p(I) \right) \sigma'_p(I, a)} \\
= & \sum_{I\in \mathcal{I}_p}\sum_{a\in A(I)}{\pi^{\sigma'}_p(I)v^{\sigma}_p(I, a) \sigma'_p(I, a)} - \\
& \sum_{I\in \mathcal{I}_p}{\pi^{\sigma'}_p(I) v^{\sigma}_p(I)}
\end{aligned}
\end{equation}

Note that $\pi^{\sigma'}_p(I)\sigma'_p(I, a) = \pi^{\sigma'}_p(h)\sigma'_p(h, a) = \pi^{\sigma'}_{p}(h') = \pi^{\sigma'}_{p}(I')$ for any $h\in I$, $h' \in Succ_p(h \cdotp a)$ and $I'\in Succ_p(I, a)$.
So, for the first term on the right side in the above equation, according to the recursive definition of counterfactual values (Equation (\ref{eq:rec_v}) in the paper),
\begin{equation}
\begin{aligned}
& \sum_{I\in \mathcal{I}_p}\sum_{a\in A(I)}{\pi^{\sigma'}_p(I)v^{\sigma}_p(I, a) \sigma'_p(I, a)} \\
= & \sum_{I\in \mathcal{I}_p}\sum_{h\in I}\sum_{a\in A(I)}{\sum_{z\in Z: z \in Succ_p(h \cdotp a)}{\pi_{p}^{\sigma'}(z)\pi_{-p}^{\sigma}(z)u_p(z)}} + \\
&\sum_{I\in \mathcal{I}_p}\sum_{a\in A(I)}\sum_{I'\in  Succ_p(I, a)}{\pi^{\sigma'}_{p}(I')v^{\sigma}_p(I')} \\
= & \sum_{z\in Z} \pi^{\sigma'}_{p}(z)\pi^{\sigma}_{-p}(z) u_p(z) + \sum_{I\in \mathcal{I}_p}\sum_{I'\in  Succ_p(I)}{\pi^{\sigma'}_{p}(I')v^{\sigma}_p(I')} \\
= & \sum_{z\in Z} \pi^{\sigma'}_{p}(z)\pi^{\sigma}_{-p}(z) u_p(z) + \sum_{I\in \mathcal{I}_p:I \neq I(\emptyset)}{\pi^{\sigma'}_{p}(I)v^{\sigma}_p(I)}.
\end{aligned}
\end{equation}
According to the definition of expected payoff,
$
\sum_{z\in Z} \pi^{\sigma'}_{p}(z)\pi^{\sigma}_{-p}(z) u_p(z) = v^{\langle\sigma'_p, \sigma_{-p}\rangle}_p
$.
So,
\begin{equation}
\begin{aligned}
& \sum_{I\in \mathcal{I}_p}\sum_{a\in A(I)}{\pi^{\sigma'}_p(I)\left(v^{\sigma}_p(I, a) - v^{\sigma}_p(I) \right) \sigma'_p(I, a)} \\
= & v^{\langle\sigma'_p, \sigma_{-p}\rangle}_p + \sum_{I\in \mathcal{I}_p:I \neq I(\emptyset)}{\pi^{\sigma'}_{p}(I)v^{\sigma}_p(I)} - \sum_{I\in \mathcal{I}_p}{\pi^{\sigma'}_p(I) v^{\sigma}_p(I)} \\
= & v^{\langle\sigma'_p, \sigma_{-p}\rangle}_p - v^{\sigma}_p(I(\emptyset)) \\
= & v^{\langle\sigma'_p, \sigma_{-p}\rangle}_p - v^{\sigma}_p.
\end{aligned}
\end{equation}
So the lemma holds.
\end{proof}

\subsection{Proof for Lemma \ref{lm:rec_v_eq_v_r}}

\begin{proof}
The proof is basically the same as the proof for Lemma \ref{lm:v_eq_v_r}, except that 
\begin{equation}
\begin{aligned}
& \sum_{I\in \mathcal{I}_p}\sum_{a\in A(I)}{\pi^{\sigma'}_p(I)v'^{\sigma}_p(I, a) \sigma'_p(I, a)} \\
= & \sum_{I\in \mathcal{I}_p}\sum_{h\in I}\sum_{a\in A(I)}{\sum_{z\in Z: z \in Succ_p(h \cdotp a)}{\pi_{p}^{\sigma'}(z)\pi_{-p}^{\sigma}(z)u_p(z)}} + \\
&\sum_{I\in \mathcal{I}_p}\sum_{I'\in  Succ_p(I)}{\pi^{\sigma'}_{p}(I')v'^{\sigma}_p(I')} \\
= & \sum_{z\in Z} \pi^{\sigma'}_{p}(z)\pi^{\sigma}_{-p}(z) u_p(z) + \sum_{I\in \mathcal{I}_p:I \neq I(\emptyset)}{\pi^{\sigma'}_{p}(I)v'^{\sigma}_p(I)}.
\end{aligned}
\end{equation}
Therefore,
\begin{equation}
\begin{aligned}
& \sum_{I\in \mathcal{I}_p}\sum_{a\in A(I)}{\pi^{\sigma'}_p(I)\left(v'^{\sigma}_p(I, a) - v'^{\sigma}_p(I) \right) \sigma'_p(I, a)} \\
= & v^{\langle\sigma'_p, {\sigma}_{-p}\rangle}_p + \sum_{I\in \mathcal{I}_p:I \neq I(\emptyset)}{\pi^{\sigma'}_{p}(I)v'^{\sigma}_p(I)} - \sum_{I\in \mathcal{I}_p}{\pi^{\sigma'}_p(I) v'^{\sigma}_p(I)} \\
= & v^{\langle\sigma'_p, {\sigma}_{-p}\rangle}_p - v'^{\sigma}_p(I(\emptyset)) \\
= & v^{\langle\sigma'_p, {\sigma}_{-p}\rangle}_p - v'^{\sigma}_p.
\end{aligned}
\end{equation}
\end{proof}

\subsection{Proof for Equation (\ref{eq:r_decomp})}
\begin{proof}
Firstly, we have 
\begin{equation}
\begin{aligned}
& R^{T + T'}_p = \max_{\sigma'_p\in \Sigma_p}\sum_{t=1}^{T + T'}v^{\langle\sigma'_p, \sigma^t_{-p}\rangle}_p - \sum_{t=1}^{T+T'}v^{\sigma^t}_p \\
= & \left(T v'^{\sigma}_p - \sum_{t=1}^T{v^{\sigma^t}_p}\right) + \\
& \max_{\sigma'_p\in \Sigma_p}\left(\sum_{t=1}^{T + T'}v^{\langle\sigma'_p, \sigma^t_{-p}\rangle}_p - T v'^{\sigma}_p - \sum_{t'=1}^{T'}v^{\sigma^{t'}}_p\right).
\end{aligned}
\end{equation}
Recall that we assume $\sigma = \overline{\sigma}^T$. So, for the second term on the right side in the above equation,
according to Lemma \ref{lm:v_eq_v_r} and Lemma \ref{lm:rec_v_eq_v_r}, we have
\begin{equation}
\label{eq:reg_sub_reg}
\begin{aligned}
& \sum_{t=1}^{T + T'}v^{\langle\sigma'_p, \sigma^t_{-p}\rangle}_p - T v'^{\sigma}_p - \sum_{t'=1}^{T'}v^{\sigma^{t'}}_p \\
= & \left(T v^{\langle\sigma'_p, \sigma_{-p}\rangle}_p - T v'^{\sigma}_p\right) +  \sum_{t'=1}^{T'}\left(v^{\langle\sigma'_p, \sigma^{t'}_{-p}\rangle}_p - v^{\sigma^{t'}}_p\right) \\
= & \sum_{I\in \mathcal{I}_p}\pi^{\sigma'}_p(I)T(v'^{\sigma}_p(I, a) - v'^{\sigma}_p(I))\sigma'_p(I, a) + \\
& \sum_{I\in \mathcal{I}_p}\pi^{\sigma'}_p(I)\sum_{t'=1}^{T'}(v^{{\sigma}^{t'}}_p(I, a) - v^{{\sigma}^{t'}}_p(I))\sigma'_p(I, a) \\
= & \sum_{I\in \mathcal{I}_p}\pi^{\sigma'}_p(I)\left(R'^T_p(I, a) + \sum_{t'=1}^{T'}r^{\sigma^{t'}}_p(I, a)\right)\sigma'_p(I, a).
\end{aligned}
\end{equation}

Note that 
\begin{equation}
\label{e:eq:sub_regret}
    R'^{T, T'}_p(I, a) = R'^T_p(I, a) + \sum_{t'=1}^{T'}r^{\sigma^{t'}}_p(I, a).
\end{equation}
So, 
\begin{equation}
\begin{aligned}
& \max_{\sigma'_p\in \Sigma_p}\sum_{I\in \mathcal{I}_p}\pi^{\sigma'}_p(I)\left(R'^T_p(I, a) + \sum_{t'=1}^{T'}r^{\sigma^{t'}}_p(I, a)\right)\sigma'_p(I, a) \\
\leq & \sum_{I\in \mathcal{I}_p} \max_a (R'^{T,T'}_p(I, a))_+,
\end{aligned}
\end{equation}
and 
\begin{equation}
\label{e:eq:r_decomp}
R^{T+T'}_p \leq  \left(T v'^{\sigma}_p - \sum_{t=1}^T{v^{\sigma^t}_p}\right) + \sum_{I\in \mathcal{I}_p} \max_a (R'^{T,T'}_p(I, a))_+.
\end{equation}
\end{proof}

\subsection{Proof for Theorem \ref{th:new_sub_bound}}

Before we prove the theorem, we would like to introduce a useful lemma. 

\begin{lemma}
\label{lm:max_le_v_le_max}
Given an arbitrary strategy $\sigma$, $T > 0$, and $v'^\sigma_p(I)$ at all infosets, compute $v'^\sigma_p(I, a)$ according to (\ref{eq:rec_sub_v}), then,
$\max_{\sigma'_p \in \Sigma_p}v^{\langle\sigma'_p, \sigma_{-p}\rangle}_p  - \frac{1}{T}\sum_{I\in \mathcal{I}_p}\max_a{(R'^{T}_p(I, a))_+} \leq v'^{\sigma}_p$. Besides, if $v'^{\sigma}_p(I) \leq \max_{a} v'^{\sigma}_p\left(I, a\right)$ at every infoset,
then, $v'^{\sigma}_p \leq \max_{\sigma'_p \in \Sigma_p}v^{\langle\sigma'_p, \sigma_{-p}\rangle}_p$.
\end{lemma}

\begin{proof}
For the first inequality, according to Lemma \ref{lm:rec_v_eq_v_r},
\begin{equation}
\begin{aligned}
& \max_{\sigma'_p \in \Sigma_p} v^{\langle\sigma'_p, \sigma_{-p}\rangle}_p - v'^{\sigma}_p   \\
= & \max_{\sigma'_p \in \Sigma_p} \sum_{I\in \mathcal{I}_p}\sum_{a\in A(I)}{\pi^{\sigma'}_p(I)\left(v'^{\sigma}_p(I, a) - v'^{\sigma}_p(I) \right)\sigma'_p(I, a)} \\
= & \max_{\sigma'_p \in \Sigma_p} \sum_{I\in \mathcal{I}_p}\sum_{a\in A(I)}{\pi^{\sigma'}_p(I)\frac{1}{T}R'^T_p(I, a) \sigma'_p(I, a)} \\
\leq & \frac{1}{T}\sum_{I\in \mathcal{I}_p}\max_a (R'^T_p(I, a))_+.
\end{aligned}
\end{equation}
Rearranging the above equation gives the result. 

For the second inequality, let $\sigma^*_p$ be the strategy that maximizes $\sum_{a\in A(I)}v'^{\sigma}_p(I, a)\sigma^*_p(I, a)$ at every infoset, i.e., $\sigma^*_p(I, a) = \mathbbm{1}_{a=\argmax v'^{\sigma}_p(I, a)}$.
According to Lemma \ref{lm:rec_v_eq_v_r},
\begin{equation}
\begin{aligned}
&   v'^{\sigma}_p  -  v^{\langle\sigma^*_p, \sigma_{-p}\rangle}_p  \\
= & \sum_{I\in \mathcal{I}_p}\sum_{a\in A(I)}{\pi^{\sigma^*}_p(I)\left(v'^{\sigma}_p(I) - v'^{\sigma}_p(I, a) \right) \sigma^*_p(I, a)} \\ 
= & \sum_{I\in \mathcal{I}_p}{\pi^{\sigma^*}_p(I)\left(v'^{\sigma}_p(I) - \max_{a\in A(I)} v'^{\sigma}_p(I, a)\right)}.
\end{aligned}
\end{equation}
When
$v'^{\sigma}_p(I) \leq \max_{a}v'^{\sigma}_p(I, a)$, we have $v'^{\sigma}_p  \leq v^{\langle\sigma^*_p, \sigma_{-p}\rangle}_p \leq \max_{\sigma'_p \in \Sigma_p} v^{\langle\sigma'_p, \sigma_{-p}\rangle}_p$.
\end{proof}

Now we can proof Theorem \ref{th:new_sub_bound}.
\begin{proof}
According to Lemma \ref{lm:max_le_v_le_max}, we have
\begin{equation}
\begin{aligned}
T v'^{\sigma}_p - \sum_{t=1}^T{v^{\sigma^t}_p} \leq & T \max_{\sigma'_p \in \Sigma_p} v^{\langle\sigma'_p, \sigma_{-p}\rangle}_p - \sum_{t=1}^T{v^{\sigma^t}_p} \\
= & \max_{\sigma'_p \in \Sigma_p} \sum_{t=1}^T v^{\langle\sigma'_p, \sigma^t_{-p}\rangle}_p - \sum_{t=1}^T{v^{\sigma^t}_p} \\
= & R^T_p.
\end{aligned}
\end{equation}
Note that $\sigma = \overline{\sigma}^T$.
So, according to (\ref{eq:r_decomp}) in the paper,
\begin{equation}
    R^{T+T'}_p \leq  R^T_p + \sum_{I\in \mathcal{I}_p} \max_a (R'^{T,T'}_p(I, a))_+.
\end{equation}
Since $\epsilon({\sigma}) = \epsilon(\overline{\sigma}^{T}) = \frac{1}{T}\sum_{p\in P}R^{T}_p$ and $\epsilon(\overline{\sigma}^{T, T'}) = \frac{1}{T + T'}\sum_{p\in P}R^{T + T'}_p$, we have
\begin{equation}
    \epsilon(\overline{\sigma}^{T, T'}) \leq \frac{T\epsilon({\sigma})}{T + T'} + \frac{\sum_{I\in \mathcal{I}_p} \max_a (R'^{T,T'}_p(I, a))_+}{T + T'}.
\end{equation}
According to (\ref{eq:sb_R_bound}) in the paper, the theorem holds.
\end{proof}

\section{Proofs for Recursive CFR}

\subsection{Proof for Proposition \ref{prop:re_eq_XFP}}
We first quote the definition of Generalized Weakened Fictitious Play (GWFP) \cite{GWFP} for completeness.
GWFP is a kind of iterative algorithm, as defined in Definition \ref{df:GWFP}. In the definition,  $\BR_{\epsilon_t}(\Pi^{t}_{-p}) \in \{\Pi'_p \in \Sigma_p: u_p(\Pi'_p, \Pi^{t}_{-p}) \geq u_p(\BR(\Pi^{t}_{-p}), \Pi^{t}_{-p})  - \epsilon_t \}$ is a $\epsilon_t$-BR against $\Pi^{t}_{-p}$.
\begin{define}
\label{df:GWFP}
\cite{GWFP}
A generalized weakened fictitious play is a process of mixed strategies, $\{\Pi^t\}, \Pi^t\in \times_{p\in P}\Sigma_p$, s.t.
\begin{equation*}
\label{eq:GWFP_ave_strategy}
\Pi^{t+1}_{p} \in (1 - \alpha_{t+1})\Pi^{t}_{p} + \alpha_{t+1}(\BR_{\epsilon_t}(\Pi^{t}_{-p}) + M_{t+1}), \forall p \in P,
\end{equation*}
with $\alpha_t \to 0$ and $\epsilon_t \to 0$ as $t \to \infty$, $\sum_{t=1}^{\infty}{\alpha_t} = \infty$, 
and $\{M_t\}$ a sequence of perturbations that satisfies $\forall \beta > 0$
\begin{equation*}
\lim_{t\to \infty}{\sup_{k}{\Big\{ \Big\|\sum_{i=t}^{k-1}{\alpha_{i+1}M_{i+1}}\Big\| s.t. \sum_{i=t}^{k-1}{\alpha_{i+1}} \leq  \beta \Big\}  }} = 0.
\end{equation*}
\end{define}

In \cite{FSP}, a special form of GWFP, named Full-width extensive-form fictitious play (XFP), is given. In XFP, $\epsilon_t$ and $M_{t}$ is set to zero and $\alpha_{t}$ is set to $\frac{1}{t+1}$ at every iteration. So $\BR_{\epsilon_t}(\Pi^{t}_{-p})$ is a best response against the mixed strategy $\Pi^t_{-p}$ of the opponent, and the mixed strategy $\Pi^t$ is the average of the best responses.

\begin{proof}
ReCFR guarantees that $v'^{\overline{\sigma}^t}_p(I) \leq \max_{a} v'^{\overline{\sigma}^t}_p\left(I, a\right)$ for all infosets.
According to Lemma \ref{lm:max_le_v_le_max},  when $\lambda^t_p(I) = 0$ at every infoset, we have $v^{\sigma^{t+1}}_p = v'^{\overline{\sigma}^t}_p = \max_{\sigma'_p}v^{\langle\sigma'_p, \overline{\sigma}^t_{-p}\rangle}_p$, i.e., $\sigma^{t+1}_p$ is a best response to $ \overline{\sigma}^t_{-p}$.
According to the definition of XFP, the proposition holds.
\end{proof}

\subsection{Proof for Proposition \ref{prop:re_eq_CFR}}
\begin{proof}
According to the definition of ReCFR, when $\lambda^t_p(I) = \sum_{a}(R^t(I, a))^2_+$ and
\begin{equation}
    t v'^{\overline{\sigma}^t}_p(I, a) = \sum_{k=1}^t v^{{\sigma}^k}_p(I, a),
\end{equation}
we have
\begin{equation}
    t v'^{\overline{\sigma}^t}_p(I) = \sum_{k=1}^t v^{{\sigma}^k}_p(I).
\end{equation}
Note that the two equations also hold when $\sum_{a}(R^t(I, a))^2_+ = 0$.
Then, according to the recursive definition of counterfactual values and RSVs, the above two equations hold at every infoset if $\lambda^t_p(I) = \sum_{a}(R^t(I, a))^2_+$ at every infoset.
Therefore, the substitute regrets recover the cumulative regrets and ReCFR recovers CFR.
\end{proof}

\subsection{Proof for Theorem \ref{th:re_bound}}

\begin{proof}
Firstly, for any $\sigma'_p \in \Sigma_p$, we have
\begin{equation}
\begin{aligned}
&  tv'^{\overline{\sigma}^t}_p - (t - 1)v'^{\overline{\sigma}^{t-1}}_p  - v^{\sigma^{t}}_p \\ 
= & (t-1) \left(v^{\langle\sigma'_p, \overline{\sigma}^{t-1}_{-p}\rangle}_p -v'^{\overline{\sigma}^{t-1}}_p\right) + \\
& \left(v^{\langle\sigma'_p, \sigma^{t}_{-p}\rangle}_p - v^{\sigma^{t}}_p\right) -  t \left(v^{\langle\sigma'_p, \overline{\sigma}^t_{-p}\rangle}_p - v'^{\overline{\sigma}^t}_p\right).
\end{aligned}
\end{equation}

According to Lemma \ref{lm:rec_v_eq_v_r}, at iteration $t > 1$, we have
\begin{equation}
\label{eq:pr_ri_bound_1}
\begin{aligned}
& t\left(v^{\langle\sigma'_p, \overline{\sigma}^t_{-p}\rangle}_p - v'^{\overline{\sigma}^t}_p\right) \\
= & t\sum_{I\in \mathcal{I}_p}\sum_{a\in A(I)}{\pi^{\sigma'}_p(I)\left(v'^{\overline{\sigma}^t}_p(I, a) - v'^{\overline{\sigma}^t}_p(I) \right) {\sigma}'_p(I, a)} \\
= & \sum_{I\in \mathcal{I}_p}\sum_{a\in A(I)}{\pi^{\sigma'}_p(I) R'^t_p(I, a) {\sigma}'_p(I, a)},
\end{aligned}
\end{equation}
and
\begin{equation}
\label{eq:pr_ri_bound_2}
\begin{aligned}
& (t-1) \left(v^{\langle\sigma'_p, \overline{\sigma}^{t-1}_{-p}\rangle}_p - v'^{\overline{\sigma}^{t-1}}_p\right) \\
= & (t -1) \sum_{I\in \mathcal{I}_p}\sum_{a\in A(I)}\pi^{\sigma'}_p(I)\\
\quad \quad & \left(v'^{\overline{\sigma}^{t-1}}_p(I, a) - v'^{\overline{\sigma}^{t-1}}_p(I) \right) {\sigma}'_p(I, a) \\
= & \sum_{I\in \mathcal{I}_p}\sum_{a\in A(I)}{\pi^{\sigma'}_p(I) R'^{t-1}_p(I, a) {\sigma}'_p(I, a)}.
\end{aligned}
\end{equation}
Besides, according to Lemma \ref{lm:v_eq_v_r},
\begin{equation}
\label{eq:pr_ri_bound_0}
\begin{aligned}
& v^{\langle\sigma'_p, \sigma^{t}_{-p}\rangle}_p - v^{\sigma^{t}}_p \\
= &  \sum_{I\in \mathcal{I}_p}\sum_{a\in A(I)}{\pi^{\sigma'}_p(I)r^{\sigma^t}_p(I, a) \sigma'_p(I, a)}.
\end{aligned}
\end{equation}
So,
\begin{equation}
\label{eq:diff_v_eq_g}
\begin{aligned}
&  tv'^{\overline{\sigma}^t}_p - (t - 1)v'^{\overline{\sigma}^{t-1}}_p  - v^{\sigma^{t}}_p \\ 
= & \sum_{I\in \mathcal{I}_p}\sum_{a\in A(I)}{\pi^{\sigma'}_p(I)g'^t_p(I, a) {\sigma}'_p(I, a)},
\end{aligned}
\end{equation}
where 
\begin{equation}
    g'^t_p(I, a) = R'^{t-1}_p(I, a) + r^{\sigma^t}_p(I, a) - R'^{t}_p(I, a).
\end{equation}
Similarly, when $t = 1$, we have
\begin{equation}
\begin{aligned}
&  v'^{\overline{\sigma}^t}_p  - v^{\sigma^{t}}_p \\ 
= & (v'^{{\sigma}^t}_p - v^{\langle\sigma'_p, \sigma^t_{-p}\rangle}_p) + (v^{\langle\sigma'_p, \sigma^t_{-p}\rangle}_p - v^{\sigma^{t}}_p) \\
= & \sum_{I\in \mathcal{I}_p}\sum_{a\in A(I)}{\pi^{\sigma'}_p(I)\left(r^{\sigma^t}_p(I, a) - R'^{t}_p(I, a)\right){\sigma}'_p(I, a)}.
\end{aligned}
\end{equation}
Let $v'^{\overline{\sigma}^0}_p = 0$ and $v'^{\overline{\sigma}^0}_p(I, a) =  v'^{\overline{\sigma}^0}_p(I) = R'^{0}_p(I, a) = 0$. Then, (\ref{eq:diff_v_eq_g}) holds for $t \geq 1$.

Let $\sigma'_p = \sigma^{t+1}_p$.
Notice that $\sigma^{t+1}_p(I,  a)  = \frac{(R'^{t}_p(I, a))_+}{\sum_{a}(R'^{t}_p(I, a))_+}$,
\begin{equation}
\begin{aligned}
& \sum_{a\in A(I)}{g'^t_p(I, a) {\sigma}^{t+1}_p(I, a)} \\
=  & \frac{\sum_{a}\left(R'^{t-1}_p(I, a) - R'^{t}_p(I, a)\right)(R'^{t}_p(I, a))_+}{{\sum_{a}(R'^{t}_p(I, a))_+}} + \\
& \frac{\sum_{a}r^{\sigma^t}_p(I, a)(R'^{t}_p(I, a))_+}{\sum_{a}(R'^{t}_p(I, a))_+} \\
\leq & \frac{\sum_{a}\left((R'^{t-1}_p(I, a))_+ - (R'^{t}_p(I, a))_+\right)(R'^{t}_p(I, a))_+}{{\sum_{a}(R'^{t}_p(I, a))_+}} + \\
& \frac{\sum_{a}r^{\sigma^t}_p(I, a)(R'^{t}_p(I, a))_+}{\sum_{a}(R'^{t}_p(I, a))_+}.
\end{aligned}
\end{equation}
Besides, we have $\sum_{a}r^{\sigma^t}_p(I, a) (R'^{t-1}_p(I, a))_+ = 0$ when $t > 1$, as $r^{\sigma^t}_p(I, a) = v^{\sigma^t}_p(I, a) - \sum_{a}v^{\sigma^t}_p(I, a)\sigma^t(I, a)$ and $\sigma^{t}_p(I,  a) \propto (R'^{t-1}_p(I, a))_+$. It is also true when $t = 1$ as $R'^{0}_p(I, a) = 0$.
So, 
\begin{equation}
\begin{aligned}
& \frac{\sum_{a}r^{\sigma^t}_p(I, a)(R'^{t}_p(I, a))_+}{\sum_{a}(R'^{t}_p(I, a))_+} \\
= & \frac{\sum_{a}r^{\sigma^t}_p(I, a)\left((R'^{t}_p(I, a))_+ - (R'^{t-1}_p(I, a))_+\right)}{\sum_{a}(R'^{t}_p(I, a))_+} \\
\leq & \frac{\sum_{a}(r^{\sigma^t}_p(I, a))^2}{2\sum_{a}(R'^{t}_p(I, a))_+} + \\
& \frac{\sum_{a}\left((R'^{t}_p(I, a))_+ - (R'^{t-1}_p(I, a))_+\right)^2}{2\sum_{a}(R'^{t}_p(I, a))_+}.
\end{aligned}
\end{equation}
The last inequality is derived according to Fenchel-Young inequality.
Combine the above two equations, we get
\begin{equation}
\begin{aligned}
& \sum_{a\in A(I)}{g'^t_p(I, a) {\sigma}^{t+1}_p(I, a)} \\
\leq & \frac{\sum_{a}\left((R'^{t-1}_p(I, a))_+ - (R'^{t}_p(I, a))_+\right)(R'^{t}_p(I, a))_+}{{\sum_{a}(R'^{t}_p(I, a))_+}} + \\
& \frac{\sum_{a}\left((R'^{t}_p(I, a))_+ - (R'^{t-1}_p(I, a))_+\right)^2}{2\sum_{a}(R'^{t}_p(I, a))_+} + \\
& \frac{\sum_{a}(r^{\sigma^t}_p(I, a))^2}{2\sum_{a}(R'^{t}_p(I, a))_+} \\
= & \frac{\sum_{a}\left(R'^{t-1}_p(I, a)\right)^2_+ - \sum_{a}\left(R'^{t}_p(I, a)\right)^2_+}{{2\sum_{a}(R'^{t}_p(I, a))_+}}+ \\
& \frac{\sum_{a}(r^{\sigma^t}_p(I, a))^2}{2\sum_{a}(R'^{t}_p(I, a))_+}.
\end{aligned}
\end{equation}
Notice that $\sum_{a}\left(R'^{t}_p(I, a)\right)^2_+ = \lambda^t_p(I)$ ((\ref{eq:re_cons}) in the paper) and $\sqrt{\sum_{a}\left(R'^{t}_p(I, a)\right)^2_+} \leq \sum_{a}(R'^{t}_p(I, a))_+$, we have
\begin{equation}
\begin{aligned}
& \frac{\sum_{a}\left(R'^{t-1}_p(I, a)\right)^2_+ - \sum_{a}\left(R'^{t}_p(I, a)\right)^2_+}{{2\sum_{a}(R'^{t}_p(I, a))_+}}+ \\
\leq & \frac{{\lambda^{t-1}_p(I)} - {\lambda^{t}_p(I)}}{2\sqrt{\lambda^{t}_p(I)}}.
\end{aligned}
\end{equation}
Then, according to (\ref{eq:diff_v_eq_g}),
\begin{equation}
\begin{aligned}
& T v'^{\overline{\sigma}^T}_p - \sum_{t=1}^T{v^{\sigma^t}_p} \\
= & \sum_{t=1}^T{\left(tv'^{\overline{\sigma}^t}_p - (t - 1)v'^{\overline{\sigma}^{t-1}}_p - v^{\sigma^t}_p\right)} \\
\leq &  \sum_{t=1}^{T}\sum_{I\in \mathcal{I}_p}\frac{\left({\lambda^{t-1}_p(I)} - {\lambda^{t}_p(I)} + \sum_{a}(r^{\sigma^t}_p(I, a))^2\right)_+}{2\sqrt{\lambda^t_p(I)}}.
\end{aligned}
\end{equation}
Finally, according to (\ref{eq:rr_decomp}) in the paper, 
\begin{equation}
\label{eq:proof_re_regret_bound}
\begin{aligned}
R^{T}_p \leq  & \left(T v'^{\overline{\sigma}^T}_p -  \sum_{t=1}^T{v^{\sigma^t}_p}\right) + \sum_{I\in \mathcal{I}_p} \max_a (R'^{T}_p(I, a))_+ \\
\leq & \sum_{t=1}^{T}\sum_{I\in \mathcal{I}_p}\frac{\left({\lambda^{t-1}_p(I)} - {\lambda^{t}_p(I)} + \sum_{a}(r^{\sigma^t}_p(I, a))^2\right)_+}{2\sqrt{\lambda^t_p(I)}} + \\
& \sum_{I\in \mathcal{I}_p}\max_a (R'^{T}_p(I, a))_+ \\
\leq & \sum_{t=1}^{T}\sum_{I\in \mathcal{I}_p}\frac{\left({\lambda^{t-1}_p(I)} - {\lambda^{t}_p(I)} + \sum_{a}(r^{\sigma^t}_p(I, a))^2\right)_+}{2\sqrt{\lambda^t_p(I)}} + \\
& \sum_{I\in \mathcal{I}_p}\sqrt{\lambda^T_p(I)}.
\end{aligned}
\end{equation}
\end{proof}

\subsection{Proof for Corollary \ref{co:re_0}}
\begin{proof}
Note that 
\begin{equation}
\label{eq:scale_r}
\begin{aligned}
{\sum_{a}(r^{\sigma^t}_p(I, a))^2} \leq & (\pi^{{\sigma}^t}_{-p}(I))^2\D^2(I)|A(I)| \\
\leq & \pi^{{\sigma}^t}_{-p}(I)\D^2(I)|A(I)|.
\end{aligned}
\end{equation}
Note that $\pi^{\overline{\sigma}^t}_{-p}(I) t = \sum_{k=1}^t\pi^{{\sigma}^k}_{-p}(I)$.
When $\lambda^t_p(I) = \pi^{\overline{\sigma}^t}_{-p}(I)\D^2(I)|A(I)|t$, we have
${\lambda^{t-1}_p(I)} - {\lambda^{t}_p(I)} + \sum_{a}(r^{\sigma^t}_p(I, a))^2 \leq 0$. 
According to Theorem \ref{th:re_bound},
\begin{equation}
\begin{aligned}
R^T_p \leq & \sum_{I\in \mathcal{I}_p}\sqrt{\lambda^T_p(I)}\\
\leq & \sum_{I\in\mathcal{I}_p}\sqrt{\pi^{\overline{\sigma}^T}_{-p}(I)}\D(I)\sqrt{|A(I)|}\sqrt{T}.
\end{aligned}
\end{equation}
As $\epsilon(\overline{\sigma}^T) =  \frac{1}{T}\sum_{p\in P}R^T_p$, the corollary holds.
\end{proof}

\subsection{Proof for Corollary \ref{co:re_1}}

\begin{lemma} 
\label{lm:sum_int}
\cite{FTRL_OMD_proof}
Let $a_t \geq 0$ for $0 \leq t < T$ and $f: [0, +\infty) \to [0, +\infty)$ a nonincresing function. Then
\begin{equation}
\sum_{t=1}^T{a_t f\left(a_0 + \sum_{k=1}^t a_k\right)} \leq \int_{a_0}^{\sum_{t=0}^T a_t}f(x)dx.
\end{equation}
\end{lemma}

\begin{proof}
\cite{FTRL_OMD_proof}.
Denote $s_t = \sum_{k=0}^t a_k$.
\begin{equation}
    a_t f\left(a_0 + \sum_{k=1}^t a_k\right) = a_t f(s_t) \leq \int_{s_{t-1}}^{s_t}f(x)dx.
\end{equation}
Summing over $t =1, \dots, T$, we have the stated bound.
\end{proof}

According to Lemma \ref{lm:sum_int}, if $\sum_{k=1}^t a^2_k > 0$ for $1 \leq t \leq T$,  it is immediately that
\begin{equation}
\sum_{t=1}^T{\frac{a^2_t}{\sqrt{\sum_{k=1}^t a^2_k}}} \leq 2\sqrt{\sum_{t=1}^T a^2_t}.
\end{equation}
Now, we can prove the corollary.

\begin{proof}
Note that $\pi^{\overline{\sigma}^t}_{-p}(I) t = \sum_{k=1}^t\pi^{{\sigma}^k}_{-p}(I)$.
When $\lambda^t_p(I) = \lambda\pi^{\overline{\sigma}^t}_{-p}(I)\D^2(I)|A(I)|t$, we have  $\lambda^{t-1}_p(I) \leq \lambda^{t}_p(I)$. According to Theorem \ref{th:re_bound}, 
\begin{equation}
\label{eq:re_bound_lambda}
\begin{aligned}
R^T_p \leq & \sum_{I\in \mathcal{I}_p}\Bigg\{\sum_{t=1}^{T}\frac{\pi^{{\sigma}^t}_{-p}(I)\D^2(I)|A(I)|}{2\sqrt{\lambda^t_p(I)}} + \sqrt{\lambda^T_p(I)}\Bigg\} \\
= & \sum_{I\in \mathcal{I}_p}\Bigg\{\sum_{t=1}^{T}\frac{\pi^{{\sigma}^t}_{-p}(I)\D^2(I)|A(I)|}{2\sqrt{\lambda\sum_{k=1}^{t}\pi^{{\sigma}^k}_{-p}(I)\D^2(I)|A(I)|}} + \\
& \sqrt{\lambda\pi^{\overline{\sigma}^T}_{-p}(I)\D^2(I)|A(I)|T}\Bigg\} \\
\leq & \left(\frac{1}{\sqrt{\lambda}} + \sqrt{\lambda}\right)\sum_{I\in\mathcal{I}_p}\sqrt{\pi^{\overline{\sigma}^T}_{-p}(I)}\D(I)\sqrt{|A(I)|}\sqrt{T}.
\end{aligned}
\end{equation}
As $\epsilon(\overline{\sigma}^T) =  \frac{1}{T}\sum_{p\in P}R^T_p$, the corollary holds.
\end{proof}

\section{Proofs for Recursive CFR with Bootstrapping}
\label{sec:conv_boot}

\subsection{Proof for Equation (\ref{eq:sb_q_exp})}
\label{sec:cfrmax_q}
\begin{proof}
The equation shows that, for any $I\in \mathcal{I}_p, a\in A(I)$,
\begin{equation}
\begin{aligned}
\label{e:eq:sb_q_exp}
u'^{\overline{\sigma}^t}_p(I, a) = &
    \mathbb{E}_{h \sim I, h'\sim Succ_p(h \cdotp a)}\big\{\mathbbm{1}_{h'\in Z}u_p(h') +\\ 
    &\mathbbm{1}_{h'\notin Z}u'^{\overline{\sigma}^t}_p(I(h'))\big\}.
\end{aligned}
\end{equation}
The right side of  (\ref{e:eq:sb_q_exp}) can be expended as
\begin{equation}
\begin{aligned}
\label{eq:sb_q_exp_expand}
&\mathbb{E}_{h \sim I, h'\sim Succ_p(h \cdotp a)}\big\{\mathbbm{1}_{h'\in Z}u_p(h') + \mathbbm{1}_{h'\notin Z}u'^{\overline{\sigma}^t}_p(I(h'))\big\} \\
= & \sum_{h\in I}{p(h|I)\sum_{h'\in Succ_p(h \cdotp a)}{p(h'|h \cdotp a)\mathbbm{1}_{h'\in Z}u_p(h') }} + \\
& \sum_{h\in I}{p(h|I)\sum_{h'\in Succ_p(h \cdotp a)}{p(h'|h \cdotp a)\mathbbm{1}_{h'\notin Z}u'^{\overline{\sigma}^t}_p(I(h')) }}, \\
\end{aligned}
\end{equation}
where $p(h|I)$ is the probability of sampling $h$ when $I$ is reached and $p(h'|h \cdotp a)$ is the probability of sampling $h'$ when $h \cdotp a$ is reached. Assume player $p$ is using strategy $\hat{\sigma}$, and player $-p$ is using the average strategy $\overline{\sigma}^t$. We have
\begin{equation}
\begin{aligned}
p(h|I) = &\frac{\pi_{p}^{\hat{\sigma}}(h)\pi_{-p}^{\overline{\sigma}^t}(h)}{\sum_{h\in I}{\pi_{p}^{\hat{\sigma}}(h)\pi_{-p}^{\overline{\sigma}^t}(h)}} \\
= &\frac{\pi_{-p}^{\overline{\sigma}^t}(h)}{\sum_{h\in I}{\pi_{-p}^{\overline{\sigma}^t}(h)}} \\
= & \frac{\pi_{-p}^{\overline{\sigma}^t}(h)}{\pi_{-p}^{\overline{\sigma}^t}(I)}.
\end{aligned}
\end{equation}
Note that $\pi_{p}^{\hat{\sigma}}(h) = \pi_{p}^{\hat{\sigma}}(I)$ for any $h\in I$. As for $p(h'|h \cdotp a)$, we have
\begin{equation}
\begin{aligned}
p(h'|h \cdotp a) & = \frac{\pi_{p}^{\hat{\sigma}}(h')\pi_{-p}^{\overline{\sigma}^t}(h')}{\pi_{p}^{\hat{\sigma}}(h \cdotp a)\pi_{-p}^{\overline{\sigma}^t}(h \cdotp a)} \\
& = \pi_{-p}^{\overline{\sigma}^t}(h \cdotp a, h')
\end{aligned}
\end{equation}
Note that $\pi_{p}^{\hat{\sigma}}(h') = \pi_{p}^{\hat{\sigma}}(h \cdotp a)$ as $h'$ is the earliest reachable history of player $p$ from $h \cdotp a$. Put them into (\ref{eq:sb_q_exp_expand}), we get 
\begin{equation}
\begin{aligned}
&\mathbb{E}_{h \sim I, h'\sim Succ_p(h \cdotp a)}\big\{\mathbbm{1}_{h'\in Z}u_p(h') + \mathbbm{1}_{h'\notin Z}u'^{\overline{\sigma}^t}_p(I(h'))\big\} \\
= & \frac{1}{\pi_{-p}^{\overline{\sigma}^t}(I)}\sum_{h\in I}{\sum_{h'\in Z: h'\in Succ_p(h \cdotp a)}{\pi_{-p}^{\overline{\sigma}^t}(h)\pi_{-p}^{\overline{\sigma}^t}(h \cdotp a, h')u_p(h') }} + \\
& \frac{1}{\pi_{-p}^{\overline{\sigma}^t}(I)}\sum_{h\in I}{\sum_{h'\notin Z: h'\in Succ_p(h \cdotp a)}{\pi_{-p}^{\overline{\sigma}^t}(h)\pi_{-p}^{\overline{\sigma}^t}(h \cdotp a, h')u'^{\overline{\sigma}^t}_p(I(h')) }}
\\= & \frac{1}{\pi_{-p}^{\overline{\sigma}^t}(I)}\sum_{h\in I}{\sum_{h'\in Z: h'\in Succ_p(h \cdotp a)}{\pi_{-p}^{\overline{\sigma}^t}(h')u_p(h') }} + \\
& \frac{1}{\pi_{-p}^{\overline{\sigma}^t}(I)}\sum_{I'\in Succ_p(I, a)}{v'^{\overline{\sigma}^t}_p(I')}, \\
= & \frac{1}{\pi_{-p}^{\overline{\sigma}^t}(I)}v'^{\overline{\sigma}^t}_p(I, a).
\end{aligned}
\end{equation}
The second equality is derived according to the the definition of $Succ_p(I, a)$. The last equality is because of the recursive property of RSVs. Because  $u'^{\overline{\sigma}^t}_p(I, a) = {v'^{\overline{\sigma}^t}_p(I, a)}/{{\pi^{\overline{\sigma}^t}_{-p}(I)}}$, the equation holds.
\end{proof}

\subsection{Proof for Theorem \ref{th:convq}}
    The Bootstrap learning at every iteration in the algorithm is a mimic of Q-learning.  According to \cite{ConvQ}, the convergence of Q-learning is guaranteed by Theorem \ref{th:convQ}.
    \begin{theorem}
    \label{th:convQ}
    \cite{ConvQ}
    The Q-learning algorithm given by 
    \begin{equation}
    \begin{aligned}
    & Q_{k+1}(s, a) =  Q_{k}(s, a) + \\
    & \alpha_{k}\left(r(s, a) + \gamma \max_{a'}{Q_{k}(s', a')} - Q_{k}(s, a)\right),
    \end{aligned}
    \end{equation}
    where $s'$ is the state transferred from $s$ after selecting action $a$,
    converges to the optimal $Q^*(s,a)$ values if
    \begin{itemize}
        \item The state and action spaces are finite.
        \item $\sum_k{\alpha_k} = \infty$ and $\sum_k{\alpha^2_k} < \infty$ uniformly w.p.1.
        \item $Var{(r(s, a))}$ is bounded.
    \end{itemize}
    \end{theorem}
    
    Similarly, we can prove Theorem \ref{th:convq} in the paper.
    
    \begin{lemma}
    \label{lm:boot_1} 
    For any ${x}\in \mathbb{R}^n$, ${y} \in \mathbb{R}^n$, $X\in \mathbb{R}$ and $Y\in \mathbb{R}$, if $\sum_{i=1}^n{(x_i - X)^2_+} = \sum_{i=1}^n{(y_i - Y)^2_+} > 0$, then, $|X - Y| \leq \max_{i}|x_i - y_i|$.
    \end{lemma}
    
    \begin{proof}
    Firstly,  for any $z\in \mathbb{R}$, we have
    \begin{equation}
    \label{eq:proof_boot_1}
    \sum_{i=1}^n{(x_i - X)^2_+} = \sum_{i=1}^n{((y_i + z) - (Y + z))^2_+}.
    \end{equation}
    Let $z = \max_{i}|x_i - y_i|$. 
    Note that $x_{i'} \leq  y_{i'} + \max_{i}|x_i - y_i|$ for any $1\leq i' \leq n$. So, we have 
    $
    X \leq Y + \max_{i}|x_i - y_i|
    $. Otherwise, (\ref{eq:proof_boot_1}) is contradicted. Similarly, we have $Y \leq X + \max_{i}|x_i - y_i|$. So, the lemma holds.
    
    \end{proof}
    
    Now, we are ready to prove Theorem \ref{th:convq} in the paper.
    
    \begin{proof}
    At infoset $I$ at iteration $t$, let a state $s$ in Theorem \ref{th:convQ} represent an infoset $I$ in Recursive CFR. Let $Q_{k+1}(s, a) = u'^{\overline{\sigma}^t,k+1}_p (I, a)$, $Q_{k}(s', a') = u'^{\overline{\sigma}^t,k}_p(I', a')$, and $r(s, a) = \mathbb{E}_{h \sim I, h'\sim Succ_p(h \cdotp a)}\mathbbm{1}_{h'\in Z}u_p(h')$.
    Following the proof in \cite{ConvQ}, we only need to prove that the following mapping operator is a contraction operator:
    \begin{equation}
    \label{eq:proof_conv_mapping}
    \begin{aligned}
    \mathbf{H} :  (\mathbf{H}u'^{\overline{\sigma}^t, k}_p)(I, a)
    =  r(I, a) +  \gamma\sum_{I'\in \mathcal{I}_p}{P^{\overline{\sigma}^t}_p(I'|I, a)u'^{\overline{\sigma}^t, k}_p(I')}
    \end{aligned}
    \end{equation}
    where $P^{\overline{\sigma}^t}_p(I'|I, a) = {\mathbbm{1}_{I'\in Succ_p(I,a)}\pi_{-p}^{\overline{\sigma}^t}(I')}/{\pi_{-p}^{\overline{\sigma}^t}(I)}$ is the probability of reach infoset $I'$ from $I$. 
    So we need to prove
    \begin{equation}
    \label{eq:proof_boot_2}
    \left\|(\mathbf{H}u'^{\overline{\sigma}^t, k}_p) - (\mathbf {H}u'^{\overline{\sigma}^t, k'}_p)\right\|_{\infty} \leq \gamma \left\|u'^{\overline{\sigma}^t, k}_p - u'^{\overline{\sigma}^t, k'}_p\right\|_{\infty},
    \end{equation}
    for any $\overline{\sigma}^t$ and $1\leq k, k' \leq K$.
    Since $r(I, a)$ is constant with respect to $k$, 
    \begin{equation}
    \begin{aligned}
    & \left\|(\mathbf{H}u'^{\overline{\sigma}^t, k}_p) - (\mathbf {H}u'^{\overline{\sigma}^t, k'}_p)\right\|_{\infty} \\
    = & \gamma\max_{I, a}\left|\sum_{I'\in \mathcal{I}_p}P^{\overline{\sigma}^t}_p(I'|I, a)\left(u'^{\overline{\sigma}^t, k}_p (I') - u'^{\overline{\sigma}^t, k'}_p (I')\right)\right|.
    \end{aligned}
    \end{equation}
    Since $P^{\overline{\sigma}^t}_p(I'|I, a) \geq 0$ and $\sum_{I'\in \mathcal{I}_p}{P^{\overline{\sigma}^t}_p(I'|I, a)} \leq 1$,
    \begin{equation}
    \begin{aligned}
    & \left\|(\mathbf{H}u'^{\overline{\sigma}^t, k}_p) - (\mathbf {H}u'^{\overline{\sigma}^t, k'}_p)\right\|_{\infty} \\
    \leq & \gamma\max_{I}\left|u'^{\overline{\sigma}^t, k}_p \left(I\right) - u'^{\overline{\sigma}^t, k'}_p \left(I\right)\right| \\
    \leq & \gamma\max_{I}\max_{a}\left|u'^{\overline{\sigma}^t, k}_p \left(I, a\right) - u'^{\overline{\sigma}^t, k'}_p \left(I, a\right)\right| \\
    = & \gamma \left\|u'^{\overline{\sigma}^t, k}_p  - u'^{\overline{\sigma}^t, k'}_p \right\|_\infty.
    \end{aligned}
    \end{equation}
    The second inequality is because of Lemma \ref{lm:boot_1}. Note that when $\lambda^t_p(I) = 0$, i.e.,  $u'^{\overline{\sigma}^t, k}_p \left(I\right) = \max_{a}u'^{\overline{\sigma}^t, k}_p \left(I, a\right)$ and  $u'^{\overline{\sigma}^t, k'}_p \left(I\right) = \max_{a}u'^{\overline{\sigma}^t, k'}_p \left(I, a\right)$, the inequality also holds.
    So, when $\gamma < 1$, $\mathbf{H}$ is a $\gamma$-contraction mapping.
    According to the updating rules of RSVs ((\ref{eq:re_cons_1}) and (\ref{eq:sb_q_exp}) in the paper), $u'^{\overline{\sigma}^t}_p(I, a)$ is a fixed point of the mapping shown in (\ref{eq:proof_conv_mapping}). So, $u'^{\overline{\sigma}^t, K}_p(I, a)$ converges to $u'^{\overline{\sigma}^t}_p(I, a)$ w.p.1 when $K\to \infty$ for any $(I, a)$.
    Moreover, when $\gamma = 1$, according to \cite{ConvQ}, $u'^{\overline{\sigma}^t, K}_p(I, a)$ can still converge, as long as all the terminal histories are visited with probabilities greater than 0. 
    Since we only consider depth-limited games, the theorem holds.
    \end{proof}

\section{Rules of Heads-up Limit Texas Hold’em and Flop Hold'em Poker}
\label{sec:rule}
Heads-up Limit Texas Hold’em (HULH) is a two-player zero-sum game.  At the beginning of the game, player 1 should place \$50 and player 2 should place \$100 on the desk. Then 2 private cards are dealt to each player and the game goes to the first round. The game contains 4 rounds. In each round, the two players take action in turn. A player can choose fold, call or raise. Action fold means the player gives up the game thus the game terminates immediately and the player loses the money on the desk. If a player chooses to call, he should place the same money to the desk as the other player, and the game goes to the next round or terminates if it is the final round. When a player chooses to raise, he should place more money on the desk than the other player. However, there can not be more than three raises in the first two rounds and more than four raises in the second two rounds. Raises in the first two rounds are \$100 and raises in the second two rounds are \$200. In the first round, player 1 should act first, while player 2 acts first in the rest of the rounds. When the first round ends, three community cards are dealt face-up on the desk and the second round starts. Another two community cards are dealt at the beginning of the next two rounds, one for each round, followed by a series of betting.
If the rounds end without any player folds, the two players should reveal their private cards and the five cards for the players (2 private cards plus three community cards) are compared. Then the money on the desk is won by the player with stronger cards or split evenly if a tie.

Flop Hold'em Poker (FHP) is a simplified HULH, which only contains the first two rounds of betting.

\begin{figure}[htpb]
    \centering
    \begin{subfigure}[b]{0.6\linewidth}
    \centering
    \resizebox{\linewidth}{!}{\input{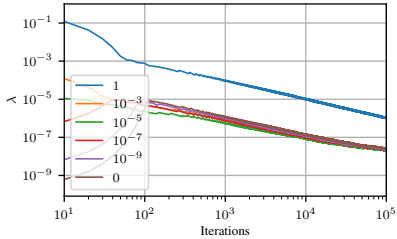}}
        \vspace*{-0.25in}
    \end{subfigure}
    \caption{Curves of the adaptive $\lambda$ in ReCFR with different initial values on Leduc poker.}
    \label{fig:lambda_leduc}
\end{figure}

\begin{figure}[htpb]
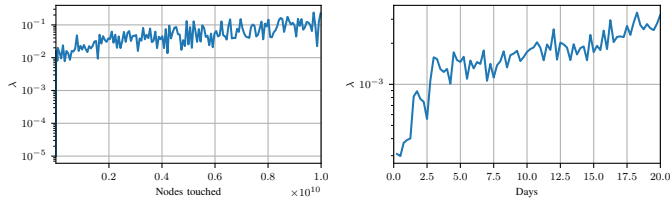

    \centering
    \begin{subfigure}[b]{0.493\linewidth}
        \centering
        \resizebox{\linewidth}{!}{\input{images/plot_alpha_0.pgf}}
        \vspace*{-0.25in}
    \end{subfigure}
    \hfill
    \begin{subfigure}[b]{0.493\linewidth}
        \centering
        \resizebox{\linewidth}{!}{\input{images/plot_lbr_alpha_0.pgf}}
        \vspace*{-0.25in}
    \end{subfigure}
    \caption{Curves of the adaptive $\lambda$ in Neural ReCFR-B on FHP (\textbf{left}) and HULH poker (\textbf{right}). }
    \label{fig:lambda}
\end{figure}

\begin{figure}[htp]
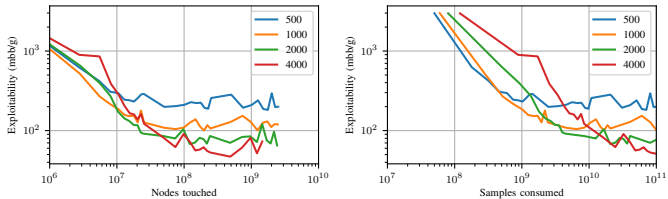

\centering
    \begin{subfigure}[b]{0.493\linewidth}
        \centering
        \resizebox{\linewidth}{!}{\input{images/plot_node_1.pgf}}
        \vspace*{-0.25in}
    \end{subfigure}
    \hfill
    \begin{subfigure}[b]{0.493\linewidth}
        \centering
        \resizebox{\linewidth}{!}{\input{images/plot_sample_1.pgf}}
        \vspace*{-0.25in}
    \end{subfigure}
     \caption{Exploitability curves of Deep CFR with different SGD steps on FHP. The x-axes represent the number of nodes touched and the number of samples consumed, respectively.}
    \label{fig:comp_deep}
\end{figure}


\begin{figure}[htbp]
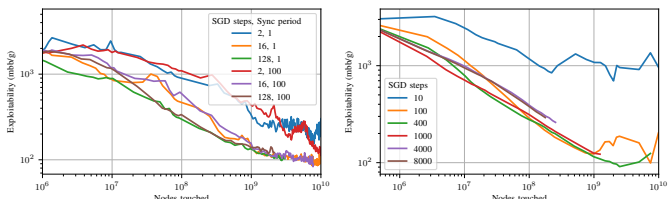

	\centering
    \begin{subfigure}[b]{0.493\linewidth}
        \centering
        \resizebox{\linewidth}{!}{\input{images/plot_node_tf_nfsp_train.pgf}}
        \vspace*{-0.25in}
        \label{fig:plot_node_tf_nfsp_train}
    \end{subfigure}
    \hfill
    \begin{subfigure}[b]{0.493\linewidth}
        \centering
        \resizebox{\linewidth}{!}{\input{images/plot_node_tf_double_cfr_runmpi_double.pgf}}
        \vspace*{-0.25in}
        \label{fig:plot_node_tf_dncfr_train}
    \end{subfigure}
    \caption{Exploitability curves of NFSP and DNCFR with different SGD steps.}
    \label{fig:NFSP_DNCFR}
\end{figure}

\section{Hyper-parameters and Experimental Environment}
\label{sec:parameter}
The hyper-parameters for all the algorithms are listed in Table \ref{tab:Parameters}. We implement Deep CFR and DREAM with the default hyper-parameters given in \cite{DeepCFR} and \cite{dream}, respectively.
The hyper-parameters for NFSP and DNCFR are decided according to a set of experiments, see Figure \ref{fig:NFSP_DNCFR}.

All the algorithms are implemented in C++ based on TensorFlow \cite{tensorflow} C++ API. All the experiments are run in a high performance computing cluster using 10 CPU cores and 40GB of memory for 10 days.
The experiments for Neural CFR-B and Deep CFR on HULH are conducted on a server with 10 CPU cores, 100GB of memory and one 2080TI GPU. The operating system is Ubuntu 18.04 and the compiler is GCC-9.0. The implementations are based on an open-source framework ``OpenSpiel''\cite{openspiel}. We seed the random generator using a real random device (std::random\_device in C++).

\section{Additional Results}

In this section, additional results of ReCFR, Neural ReCFR-B, and other algorithms are given. 

In Figure \ref{fig:lambda_leduc} and \ref{fig:lambda}, the curves of the adaptive $\lambda$ in ReCFR and Neural ReCFR-B are given. As we can see, the $\lambda$ will generally converge, and it converges to different values for different games.

In Figure \ref{fig:comp_deep}, the results of Deep CFR with different SGD steps are given. As we can see, Deep CFRs with fewer SGD steps converge earlier to higher exploitability.
The results of DNCFR and NFSP on FHP are given in Figure \ref{fig:NFSP_DNCFR}.

\begin{table*}[!htbp]
\centering
\caption{Hyper-parameters}\label{tab:Parameters} 
\begin{tabular}{llc}
\toprule
Algorithm & Hyper-parameters& \\
\midrule
Deep CFR& \makecell[l]{
sampling method = external sampling CFR \cite{MCCFR},\\ 
traversals per iteration = 10,000,\\
regret memory size = 40 million,\\
strategy memory size = 40 million,\\
optimizer = Adam,\\
learning rate = 0.001,\\
batch size = 10,000,\\
SGD steps per training for regret network = 4,000,\\
SGD steps per training for strategy network = 4,000,\\
training regret network from scratch = true,\\
training strategy network from scratch = false.
} & \\
\midrule
Deep OSCFR &
\makecell[l]{
Same as Deep CFR except \\
sampling method = outcome sampling CFR \cite{MCCFR}, \\
traversals per iteration = 50,000.\\
} & \\
\midrule
Deep CFR on HULH &
\makecell[l]{
Same as Deep CFR except \\ 
batch size = 20,000,\\
SGD steps per training for regret network = 32,000,\\
SGD steps per training for strategy network = 32,000.\\
} & \\
\midrule
DREAM &
\makecell[l]{
Same as Deep CFR except \\
sampling method = outcome sampling CFR \cite{MCCFR}, \\
traversals per iteration = 50,000,\\
SGD steps per training for regret network = 3,000,\\
SGD steps per training for strategy network = 3,000,\\
global value memory size = 200,000,\\
batch size for global value network = 512, \\
SGD steps per training for global value network = 1,000.\\
} & \\
\midrule
DNCFR&
\makecell[l]{
sampling method = CFR+ with robust sampling \cite{DoubleCFR}, \\
traversals per iteration = 10,000,\\
regret memory = none,\\
strategy memory = none,\\
optimizer = Adam,\\
learning rate = 0.001,\\
batch size = 10,000,\\
SGD steps per training for regret network = 400,\\
SGD steps per training for strategy network = 400,\\
training regret network from scratch = false,\\
training strategy network from scratch = false.\\
} & \\
\midrule
NFSP & \makecell[l]{
sampling method = self-play (trajectory sampling),\\ 
plays per iteration = 1,000,\\
value memory size = 1 million,\\
strategy memory size = 10 million,\\
optimizer = Adam,\\
learning rate = 0.001,\\
batch size = 128,\\
SGD steps per training for value network = 16,\\
SGD steps per training for strategy network = 16,\\
training value network from scratch = false,\\
training strategy network from scratch = false,\\
anticipatory parameter = 0.1.} & \\
\midrule
Neural CFR-B on HULH & \makecell[l]{
Same as NFSP except \\
RSV memory size = none (default) / 1 million (with memory),\\
SGD steps per training for RSV network = 2 epochs (default) / 32 (with memory),\\
training RSV network from scratch = false,\\
$\lambda_{init} = 10^{-5}$, \\
$\beta_{amp}  = 1.01$, \\
$\beta_{damp}  = 0.99$. \\
} & \\
\midrule
Neural CFR-B on HULH & \makecell[l]{
Same as Neural CFR-B except \\ 
plays per iteration = 100,000,\\
RSV memory size = 4 million,\\
strategy memory size = 40 million,\\
batch size = 6400,\\
SGD steps per training for RSV network = 64,\\
SGD steps per training for strategy network = 64,\\
$\lambda_{init} = 10^{-4}$. \\
} & \\
\bottomrule
\end{tabular}
\end{table*}

\end{document}